\documentclass{article}

\usepackage{arxiv}

\usepackage[utf8]{inputenc} 
\usepackage[T1]{fontenc}    
\usepackage{hyperref}       
\usepackage{url}            
\usepackage{booktabs}       
\usepackage{amsfonts}       
\usepackage{nicefrac}       
\usepackage{microtype}      
\usepackage{lipsum}         
\usepackage{graphicx}
\usepackage[numbers]{natbib}
\usepackage{doi}

\usepackage[table,xcdraw]{xcolor}
\usepackage{subfigure}
\usepackage[normalem]{ulem}
\usepackage{array}
\usepackage{caption} 
 \usepackage{multirow} 
 \usepackage{amsmath}
\usepackage{amssymb}
\usepackage{mathtools}
\usepackage{amsthm}
\usepackage{cleveref}       

\usepackage{bm}

\title{TSRM: A Lightweight Temporal Feature Encoding Architecture for Time Series Forecasting and Imputation}

\date{}

\newif\ifuniqueAffiliation

\ifuniqueAffiliation 

\else
\usepackage{authblk}

\newbox{\orcid}\sbox{\orcid}{} 
\author[1]{%
	{\usebox{\orcid}\hspace{1mm}Robert Leppich}%
}
\author[1]{%
	{\usebox{\orcid}\hspace{1mm}Michael Stenger}%
}
\author[1]{%
	{\usebox{\orcid}\hspace{1mm}Daniel Grillmeyer}%
}
\author[1]{%
	{\usebox{\orcid}\hspace{1mm}Vanessa Borst}%
}
\author[1]{%
	{\usebox{\orcid}\hspace{1mm}Samuel Kounev}%
}
\affil[1]{Department of Computer Science, University of Wuerzburg, Germany}
\fi

\renewcommand{\headeright}{PREPRINT}
\renewcommand{\shorttitle}{Time Series Representation Models}

\hypersetup{
pdftitle={TSRM: A Lightweight Temporal Feature Encoding Architecture for Time Series Forecasting and Imputation},
}

\def\eqref#1{equation~\ref{#1}}

\def\1{\bm{1}}

\def\ve{{\bm{e}}}

\def\vm{{\bm{m}}}

\def\vr{{\bm{r}}}

\def\vx{{\bm{x}}}
\def\vy{{\bm{y}}}

\def\mE{{\bm{E}}}

\def\mR{{\bm{R}}}

\DeclareMathAlphabet{\mathsfit}{\encodingdefault}{\sfdefault}{m}{sl}
\SetMathAlphabet{\mathsfit}{bold}{\encodingdefault}{\sfdefault}{bx}{n}

\newcommand{\EncodingLayer}{\textit{EL}}
\newcommand{\RepresentationLayer}{\textit{RL}}
\newcommand{\MergeLayer}{\textit{ML}}

\begin{document}
\maketitle

\begin{abstract}
	We introduce a temporal feature encoding architecture called Time Series Representation Model (TSRM) for multivariate time series forecasting and imputation. The architecture is structured around CNN-based representation layers, each dedicated to an independent representation learning task and designed to capture diverse temporal patterns, followed by an attention-based feature extraction layer and a merge layer, designed to aggregate extracted features. The architecture is fundamentally based on a configuration that is inspired by a Transformer encoder, with self-attention mechanisms at its core. The TSRM architecture outperforms state-of-the-art approaches on most of the seven established benchmark datasets considered in our empirical evaluation for both forecasting and imputation tasks. At the same time, it significantly reduces complexity in the form of learnable parameters. The source code is available at \url{https://github.com/RobertLeppich/TSRM}.
\end{abstract}


\section{Introduction}
\label{sec:intro}

Time series analysis has high potential in both science and industry. It comprises various disciplines, including time series forecasting, classification, and imputation. By analyzing time series data, we can gain deeper insights into various systems, such as sensor networks~\citep{papadimitriou2006optimal}, finance~\citep{zhu2002statstream}, and biological systems like the human body~\citep{ek2023transformer}. Time series forecasting plays a vital role in predicting future trends and supporting data-driven decision-making across domains such as finance, healthcare, and supply chain management, facilitating improved resource allocation and risk mitigation. In contrast, time series imputation is essential for handling missing data, particularly in sensor-generated time series, ensuring data integrity and reliability for downstream analysis.

Time series (TS) data often exhibit high dimensionality, with relationships between data points governed by both temporal dependencies and attribute-level structure. Typically, TS are recorded continuously, capturing one observation of one or multiple values at each time step. As single observations usually lack sufficient semantic information for in-depth analysis, research primarily emphasizes temporal variations. These variations offer richer insights into the intrinsic properties of TS, such as continuity and intricate temporal patterns. Since multiple overlapping variations can exist simultaneously, such modeling of temporal dynamics is particularly challenging.


Despite advancements in methodologies to tackle those challenges, such as Recurrent Neural Networks~(RNNs) and Convolutional Neural Networks~(CNNs), issues like the curse of dimensionality and vanishing/exploding gradients persist, restricting the information flow over long sequences as observed by~\cite{hochreiter2001gradient}.
With the work of~\cite{vaswani2017attention}, the Transformer architecture was proposed, and it was soon applied in the domain of TS analysis~\citep{wu2020deep}. However, due to its design for Natural Language Processing~(NLP) and the resulting use of the point-wise attention mechanism on word embeddings, approaches based on this architecture could not adequately capture all relevant TS characteristics, and they suffered from high computational and memory demands with long-term sequences~\citep{huang2018improved, povey2018time}. With time, more specialized implementations emerged. TS forecasting started with improvements to the basic Transformer architecture to overcome the memory bottleneck~\citep{li2019enhancing} with sparse-attention concepts, followed by various enhancements, especially further modifications to the attention mechanism. Examples include Informer~\citep{zhou2021informer}, Autoformer~\citep{wu2021autoformer}, and FEDFormer~\citep{zhou2022fedformer}. 
Transformer-based TS imputation has evolved around hybrid combinations of transformer components with CNN, RNN, auto-encoder, or GAN concepts~\citep{cao2018brits, fortuin2020gp, luo2018multivariate}, with recent successes such as SAITS~\citep{du2023saits}. Despite these sophisticated approaches, the recent work of~\cite{zeng2023transformers} presented a simple linear model that outperforms all previous models on a number of benchmarks, thus fundamentally questioning the use of transformer models for time series analysis. 

Novel approaches have addressed this challenge by abstracting the input data to exploit the modeling capability of transformers more effectively. Most noteworthy approaches evolved around the concept of patching, where the input sequence is split into subsequences~\citep{nie2022time, liu2024itransformer, chen2024pathformer}. To improve the modeling of temporal variations, other approaches capture temporal patterns at different abstraction levels to learn hierarchical representations of TS~\citep{wu2022timesnet}.
We provide a detailed related work section and delimitation to our work in Appendix~\ref{sec:rw}.
Note that we refer to the learning of temporal representations to extract meaningful patterns and dependencies of TS as \emph{representation learning}.

In this paper, we extend the concept of hierarchical learning of temporal representations by proposing a lightweight and adaptive multidimensional framework that supports high configurability to handle complex temporal variations and be applicable to many datasets. Our proposed architecture integrates a stackable multilayered structure for temporal representation learning, called encoding layer (\EncodingLayer{}). In each \EncodingLayer{}, a representation layer learns a distinct temporal representation of the input series. Crucially, our \EncodingLayer{} structures are designed to be stackable, supporting independent capture of representations at a different hierarchical level, restoring the original input dimensions to allow for hierarchical stacking of layers independent of the input dimension.

We summarize the key contributions of this work as follows:

\begin{itemize}
    \item We introduce Time Series Representation Models (TSRM), a novel architecture for time series analysis that integrates self-attention mechanisms with hierarchical representation learning. Unlike existing models that rely on deep or wide architectures with millions of parameters, TSRM employs an efficient multi-scale decomposition strategy, reducing computational complexity while preserving expressiveness.
    \item We conduct an extensive empirical evaluation on seven benchmark datasets for time series forecasting and imputation, comparing TSRM against state-of-the-art (SOTA) models, including transformer-based approaches. TSRM outperforms all SOTA models in forecasting for all datasets, and achieves comparable or superior performance in imputation on the majority of datasets, demonstrating its robustness and efficiency across diverse data distributions.
    \item TSRM is designed with interpretability and explainability in mind, leveraging hierarchical representations to provide insights into learned temporal structures. This facilitates model diagnosis, optimization, and adaptation, making TSRM more transparent and adaptable than existing black-box deep learning models.
    \item To foster reproducibility and further research, we publicly release the TSRM codebase, enabling researchers to benchmark and extend our work.
\end{itemize}

\section{Methodology}\label{sec:methodology}

To address the above challenges in time series analysis, we propose a Time Series Representation Model (TSRM), which provides a modular and dynamic architecture to model temporal patterns derived from different periods while keeping a low memory profile, including a remarkably small number of trainable parameters. TSRM consists of multiple stackable encoding layers~(\EncodingLayer{}), each equipped with a learnable multidimensional representation learning process to capture multiple temporal variations within the input data. Similar to PatchTST~\citep{nie2022time}, we explore a channel-independent approach, where each feature channel is processed independently using the same TSRM backbone. However, recognizing that some time series depend on correlations between features, we also introduce an alternative version of TSRM, called TSRM\_{IFC}, which goes beyond the channel-independent approach to facilitate the learning of \textbf{i}nter-\textbf{f}eature \textbf{c}orrelations. Details of this second architectural variant are provided later in this section.

\subsection{Model Architecture}\label{section:model_architecture}
\begin{figure*}
\centering
\includegraphics[width=.8\textwidth]{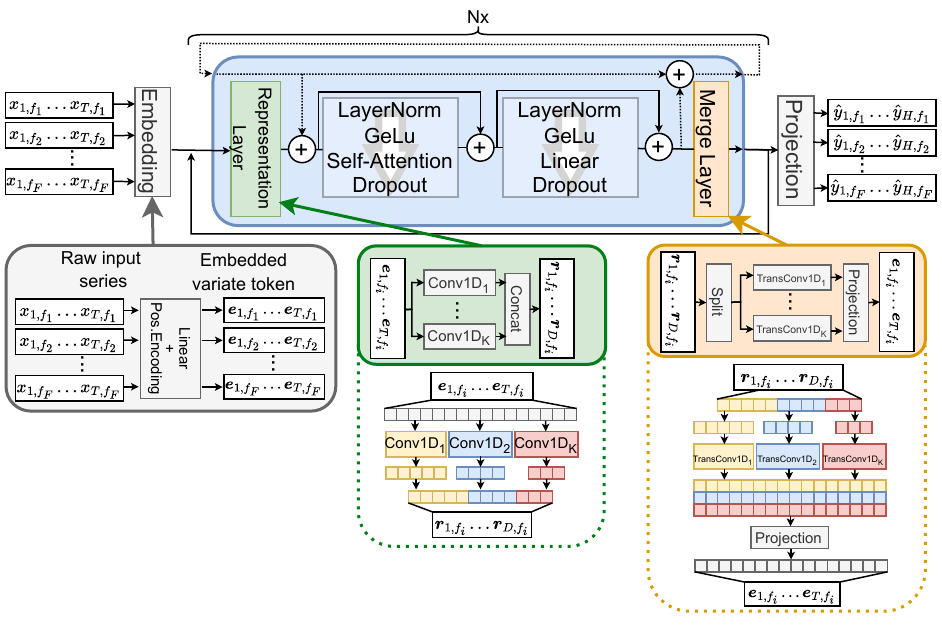}

\caption{Illustration of the proposed Time Series Representation Model (TSRM) framework, primarily composed of $N$ encoding layers (\EncodingLayer{}s) (upper section in blue), accompanied by the representation layer (\RepresentationLayer{}) (left, in green) and merge layer (\MergeLayer{}) (right, in orange).}
\label{figure:app:architecture_all}
\end{figure*}

The multivariate input sequence $\vx_1 \dots \vx_T$, where $\vx_i \in \mathbb{R}^F$ and $F$ is the amount of input features, is split into $F$ univariate sequences, where each of them is fed independently into the model according to our channel-independent setting. Notably, all of them share the same TSRM backbone, as illustrated in Figure~\ref{figure:app:architecture_all}. 

Each univariate input sequence $x_{1,f_i} \dots x_{T,f_i}$, where $i \in \{1, ..., F\}$, undergoes a position-wise operation, which extends the univariate sequence to the dimension $d$ and is then added to a positional embedding and normalized with RevIN~\citep{kim2021reversible}, resulting in the embedded sequence $\ve_{1,f_i} \dots \ve_{T,f_i}$, where $\ve_{*,f_i} \in \mathbb{R}^d$, as illustrated in Figure~\ref{figure:app:architecture_all} (left, in grey).
Subsequently, the embedded sequence undergoes processing via $N$ consecutive \EncodingLayer{}s, each tasked with deriving representations, learning and encoding temporal features, aggregating encoded features, and restoring the input dimension. These layers utilize the sequence received from the input TS or the preceding \EncodingLayer{}, enabling a hierarchical representation learning. The output of each \EncodingLayer{} is fed into the next, as well as a residual connection, marked with dotted lines in Figure~\ref{figure:app:architecture_all}, bridging representation matrices across the \EncodingLayer{}s. This stacking of layers and the residual connections facilitate a structured feature extraction, similar to deep CNN frameworks known from computer vision~\citep{he2016deep}, and they follow the information-flow principles from~\cite{hochreiter2001gradient}. Following $N$ \EncodingLayer{}s, we utilize a feed-forward layer to compute the output sequence, based on the concept proven by~\cite{das2023long}. Finally, we denormalize and reshape the sequence to its original multivariate representation.
In the following, we describe each component in detail.

\paragraph{Encoding Layer (\EncodingLayer{})}
Unlike information-rich word embeddings in NLP, which help models learn language patterns effectively~\citep{selva2021review}, the informational value of one TS observation at a single point in time is naturally lower, only gaining context information when combined across time steps or the feature dimension.
To mitigate this challenge, we introduce the self-attention-based \EncodingLayer{}, which utilizes a potent representation methodology capable of efficiently encapsulating both local and global contextual information. Furthermore, our approach is based on multi-level representation learning, where each \EncodingLayer{} learns a representation of the input sequence, highlights essential patterns, and returns the representation sequence with the dimension of the input, aggregating and embedding essential information. The \mbox{\EncodingLayer{},} presented in Figure~\ref{figure:app:architecture_all} (top in blue), is structured as follows:
It starts with the representation layer (\RepresentationLayer{}), tasked with acquiring representations embedded within the sequence, and concludes with the merge layer (\MergeLayer{}), which aggregates all acquired representations and restores the original input dimension. Hence, the \EncodingLayer{}s are structured to maintain the dimensionality of the input at the output, allowing for $N$ (i.e., the number of \EncodingLayer{} blocks) to be modulated as an independent hyperparameter, which is unrelated to the initial sequence length. Between both layers, there are two blocks of layers which are comparable to the Transformer Encoder and have self-attention and linear transformation at their core.
Below, we explain in more detail each of the two novel layers as well as the two layer blocks in-between.

\paragraph{Representation Layer (\RepresentationLayer{})}
The representation layer (\RepresentationLayer{}) is designed to independently learn representations of different abstraction levels from an input sequence as depicted in Figure~\ref{figure:app:architecture_all} (bottom in green). Unlike previous approaches~\citep{nie2022time, liu2024itransformer}, which rely on static patches, our method employs a setup of $K$ independent 1D CNN layers with varying kernel sizes, which are designed to capture and integrate representations across different abstraction levels. 
Some of the $K$ CNN layers employ small kernels without dilation for identifying basic features (such as sequence details), while others use medium-sized kernels with minimal dilation for intermediate feature recognition, or large kernels with significant dilation for detecting comprehensive features like trends. 

To enable a higher level of abstraction, we employ dilation in larger kernels. By default, all kernels are configured with a stride equal to the kernel size to limit the memory footprint.
The number of individual CNN layers~($K$), as well as the kernel size and dilation, are hyperparameters and are thus subject to hyperparameter study to find the right constellation for a specific dataset.
The outcomes from the $K$ CNN layers are concatenated on the sequence dimension, effectively transforming the input sequence $\ve_{1,f_i} \dots \ve_{T,f_i}$ into the representation $\vr_{1,f_i} \dots \vr_{D,f_i}$, where $D$ corresponds to the total length of the concatenated matrices. 
This aggregation encapsulates the encoded feature information for each feature $f_i$, spanning varied abstraction levels. The process is comparable to the representation learning of TimesNet~\citep{wu2022timesnet}. However, instead of FFT-based capturing and embedding different abstractions into a two-dimensional encoding, we use efficient one-dimensional CNN layers with different abstraction levels and smaller embeddings to enable a low memory profile and fewer trainable parameters, thus reducing the model complexity. The process of the \RepresentationLayer{} for each feature $f_i$ is formalized in Eq.~\ref{eq:representation_layer}. The index $f_i$ is omitted for clarity. $\delta_j$ and $s_j$ denote the dilation and kernel size of the $j$-th Conv1D layer with its stride equal to the kernel size and without padding.

\begin{equation}
\label{eq:representation_layer}
\begin{aligned}
    \mE &:= \ve_1 \dots \ve_T, &\mE &\in \mathbb{R}^{T \times d}\\
    \mR_j &= \text{Conv1D}_j(\mE), &\mR_j &\in \mathbb{R}^{D_j \times d},\\
    & &j&\in \{1, \dots, K\}\\
    D_j &= \Bigl\lfloor\frac{T-\delta_j (s_j-1)-1}{s_j} + 1\Bigr\rfloor\\
    \mR &= \text{Concat}(\mR_1, \dots, \mR_K), &\mR &\in \mathbb{R}^{D \times d},\\
    & &D &= \textstyle\sum_{j=1}^{K}{D_j}\\
    &\vr_1 \dots \vr_D := \mR
\end{aligned}
\end{equation}

\paragraph{Merge Layer}
The merge layer (\MergeLayer{}), illustrated in Figure~\ref{figure:app:architecture_all} (bottom right in orange), is designed to reverse the dimensional alterations caused by the \RepresentationLayer{} and to aggregate the discovered representations. It comprises $K$ 1D transposed convolution layers and utilizes transposed kernels to invert the transformations applied by the corresponding 1D CNN layer of the \RepresentationLayer{}, thereby reinstating the original data dimensions.
Therefore, for each feature, the sequence $\vr_{1,f_i} \dots \vr_{D,f_i}$ is segmented in the opposite way compared to the concatenation in the \RepresentationLayer{}.
These $K$ matrices are then merged using a feed-forward projection, resulting in a sequence that has the exact dimensions as the original input sequence for the \RepresentationLayer{}, $\ve_{1,f_i} \dots \ve_{T,f_i}$. The process of the \MergeLayer{} for each feature $f_i$ is formalized in Eq.~\ref{eq:merge_layer}. The index $f_i$ is omitted for clarity. The resulting dimensions $D_j$ of the split operation are the same as specified by the 1D CNN layers in the \RepresentationLayer{} in Eq.~\ref{eq:representation_layer}. Since the \MergeLayer{} has more of a restructuring and aggregating function, the gradients of the \MergeLayer{} can be deactivated and are examined as part of the hyperparameter study.

\begin{equation}
\label{eq:merge_layer}
\begin{aligned}
    \mR_1, &\dots, \mR_K := \text{Split}(\mR), &\mR &\in \mathbb{R}^{D \times d},\\
    & &\mR_j &\in \mathbb{R}^{D_j \times d},\\
    & &j &\in \{1, \dots, K\}\\
    \mR_j' &= \text{TransConv1D}_j(\mR_j), &\mR_j' &\in \mathbb{R}^{T \times d}\\
    \mR' &= \text{Concat}(\mR_1', \dots, \mR_K'), &\mR' &\in \mathbb{R}^{T \times dK}\\
    \mE &= \text{FeedForward}_{dK \times d}(\mR'), &\mE &\in \mathbb{R}^{T \times d}\\
    \ve_1 \dots \ve_T &:= \mE
\end{aligned}
\end{equation}

\paragraph{Middle Encoding Blocks}
Situated between the \RepresentationLayer{} and the \MergeLayer{}, two blocks of layers facilitate the extraction and amplification of features from the \RepresentationLayer{}. Each is encapsulated within a residual connection following the pre-activation design paradigm~\citep{he2016identity}. The initial block starts with a layer normalization, which is succeeded by a GeLu activation function, a multi-head self-attention mechanism, and a dropout operation. The implementation of the multi-head self-attention spans conventional attention akin to the Transformer and sparse self-attention mimicking~\cite{wu2020adversarial}. The choice between these two attention mechanisms is adjustable via hyperparameters. 

The sequence then advances to the second block, initiating again with a layer normalization, followed by a GeLu activation, a linear layer, and a dropout. The linear layer embedded within this block differs between the two architecture versions, TSRM and TSRM\_{IFC} (Inter Feature Correlation). In TSRM, the linear layer connects all $d$ dimensions, preserving channel independence. In the TSRM\_{IFC}, however, it spans all $F \times d$ dimensions, thereby facilitating the learning of inter-feature correlations.

Residual connections interlink all of the learned representations across the \EncodingLayer{}s. These connections are added to the resultant matrix from the \RepresentationLayer{}. In addition, the attention and feature-correlation augmented matrix is fused with the residual connection prior to the introduction of the \MergeLayer{}, aimed to foster information propagation independent of the \RepresentationLayer{} and the \MergeLayer{}.

\paragraph{Explainability}
Due to the lean and less complex architecture design, it is possible to transfer all attention weights of the \EncodingLayer{}s back and map them to the input time series. Hence, engineers are able to understand the weighting of the attention layer for each of the $N$ \EncodingLayer{}s and for each feature separately. This, in turn, permits insight into the functionality and effectiveness of the representation learning process, providing for a certain degree of explainability regarding the model intrinsics. This not only proves extremely useful during training on a new dataset but also provides valuable insights about essential patterns in the input sequence for the corresponding task, such as forecasting. We provide a detailed example in Appendix~\ref{appendix:exp_explain1}.
\section{Experiments}

In order to assess the effectiveness of our proposed architecture, we conducted a series of experiments using publicly accessible and well-established benchmark datasets from different fields: ECL~\citep{dua2017uci}, ETT~\citep{zhou2021informer} (four subsets: ETTm1, ETTm2, ETTh1, ETTh2), Weather and Exchange~\citep{wu2021autoformer}. All datasets were collected from~\cite{wu2021autoformer}. For more details, we refer to Section~\ref{appendix:datasets}.

\subsection{Experimental Setup}\label{sec:exp_setup}

For all experiments, we employed early stopping with a threshold of 1\% performance increase on the mean squared error (MSE) metric during validation with a patience of three epochs. Hyperparameter tuning was conducted through a random search encompassing various parameters: count of stacked \mbox{\EncodingLayer{}s} $N$, number of attention heads $h$, encoding dimension $d$, type of attention mechanism (vanilla and sparse; see Section~\ref{section:model_architecture}), activated \MergeLayer{} gradients, and configurations of the \RepresentationLayer{}, including the number of CNN layers, kernel dimensions, dilations, and grouping (see Appendix~\ref{app:exp_details} for details).
Learning rates were initially determined with an automated range test from \textit{LightningAI}~(lightning.ai) and dynamically adapted during training using a learning rate scheduler (ReduceLROnPlateau) from \textit{PyTorch} with a patience of two epochs. All models were trained with the Adam optimizer on an Nvidia A100 80GB GPU.

\subsection{Long-Term Time Series Forecasting}\label{sec:exp:fc}

Given a multivariate input sequence $\vx_{1} \dots \vx_{T}$, where $\vx_{i} \in \mathbb{R}^F$ and $F$ is the number of features, the goal is to forecast the prediction sequence $\vy_{1} \dots \vy_{H}$, where $\vy_i = \vx_{i+T}$ and $H$ denotes the prediction horizon.

To measure the discrepancy between the prediction sequence $\hat{\vy}_{1} \dots \hat{\vy}_{H}$ and the ground truth $\vy_{1} \dots \vy_{H}$ for a horizon $H$, where $\hat{\vy}_{i}, \vy_{i} \in \mathbb{R}^F$, we used the sum of the mean average error (MAE) and mean squared error (MSE) as loss function during training. 
The loss is calculated and averaged across all $F$ channels and $H$ timesteps to obtain the overall objective loss: 

\begin{equation}
    \label{eq:mse_loss}
    \mathcal{L}_\text{MAE+MSE} = 
    \frac{1}{FH}
    \sum^{H}_{i=1}{
    ||(\hat{\vy}_{i}-\vy_{i})||_{1} +
    ||(\hat{\vy}_{i}-\vy_{i})||^{2}_{2}
    }
\end{equation}

To evaluate the performance of our architecture for long-term TS forecasting, we adopted the procedure by~\cite{liu2024itransformer}:
To support a fair comparison with other approaches, we maintain the input length for all approaches at $T=96$, while varying the prediction horizon $H \in \{96, 192, 336, 720\}$. In order to increase comparability with existing work, for example, PatchTST~\cite{nie2022time}, we also provide the results of the forecasting experiments that consider $T$ as a hyperparameter in Appendix~\ref{app:ablation}. For the evaluation, we consider seven datasets (i.e., ECL, ETTm1, ETTm2, ETTh1, ETTh2, Weather, and Exchange) and compare against multiple state-of-the-art (SOTA) forecasting techniques: iTransformer~\citep{liu2024itransformer}, RLinear~\citep{li2023revisiting}, PatchTST~\citep{nie2022time}, TimesNet~\citep{wu2022timesnet} and FEDformer~\citep{zhou2022fedformer}.

\paragraph{Results}
The results, presented in Table~\ref{tab:forecasting1-avg}, show solid performance across all datasets that matches or outperforms the SOTA approaches. It can also be seen that some datasets, such as Weather, Exchange and ETTm2, achieve better results with TSRM\_IFC than with TSRM. This could be because, in these dataset, correlations between individual features are crucial for the prediction, and thus, the TSRM\_IFC model performs better, as it supports inter-feature learning. We provide further details, including a break down into different prediction horizons and a comparison against additional SOTA approaches, in Appendix~\ref{app:fc:results} and Table~\ref{app:tab:forecasting1}. The reported SOTA results, presented in Tables~\ref{tab:forecasting1-avg}~and~\ref{app:tab:forecasting1}, were taken from~\cite{liu2024itransformer}.

\begin{table*}
\centering
    \caption{Performance comparison for the multivariate forecasting task with prediction horizons $H \in \{96, 192, 336, 720\}$ and fixed lookback window $T = 96$. Results are averaged over all prediction horizons. Bold/underline indicate best/second.}
\label{tab:forecasting1-avg}
\scalebox{.88}{
\begin{tabular}{@{}ccccccccccccccc@{}}
\toprule
Models & \multicolumn{2}{c}{TSRM} & \multicolumn{2}{c}{TSRM\_M} & \multicolumn{2}{c}{iTransformer} & \multicolumn{2}{c}{RLinear} & \multicolumn{2}{c}{PatchTST} & \multicolumn{2}{c}{TimesNet} & \multicolumn{2}{c}{FEDformer} \\ \midrule
\multicolumn{1}{c|}{Metrics} & MSE & \multicolumn{1}{c|}{MAE} & MSE & \multicolumn{1}{c|}{MAE} & MSE & MAE & MSE & MAE & MSE & MAE & MSE & MAE & MSE & MAE \\ \cmidrule(l){2-15} 
\multicolumn{1}{c|}{ECL} & 0.192 & \multicolumn{1}{c|}{0.277} & \textbf{0.175} & \multicolumn{1}{c|}{\textbf{0.270}} & \underline0.178 & \textbf{0.270} & 0.218 & 0.298 & 0.205 & 0.290 & 0.192 & 0.295 & 0.214 & 0.327 \\
\multicolumn{1}{c|}{ETTm1} & \textbf{0.380} & \multicolumn{1}{c|}{\textbf{0.391}} & \underline0.384 & \multicolumn{1}{c|}{\underline0.394} & 0.407 & 0.410 & 0.414 & 0.408 & 0.387 & 0.400 & 0.400 & 0.406 & 0.448 & 0.452 \\
\multicolumn{1}{c|}{ETTm2} & \underline0.278 & \multicolumn{1}{c|}{\underline0.324} & \textbf{0.276} & \multicolumn{1}{c|}{\textbf{0.321}} & 0.288 & 0.332 & 0.286 & 0.327 & 0.281 & 0.326 & 0.291 & 0.332 & 0.304 & 0.349 \\
\multicolumn{1}{c|}{ETTh1} & \textbf{0.435} & \multicolumn{1}{c|}{\textbf{0.430}} & \underline0.440 & \multicolumn{1}{c|}{0.438} & 0.454 & 0.448 & 0.446 & \underline0.434 & 0.469 & 0.454 & 0.458 & 0.450 & \underline0.440 & 0.460 \\
\multicolumn{1}{c|}{ETTh2} & \textbf{0.372} & \multicolumn{1}{c|}{\textbf{0.395}} & 0.376 & \multicolumn{1}{c|}{0.403} & 0.383 & 0.406 & \underline0.374 & \underline0.398 & 0.387 & 0.407 & 0.414 & 0.427 & 0.436 & 0.449 \\
\multicolumn{1}{c|}{Weather} & \underline0.243 & \multicolumn{1}{c|}{\underline0.268} & \textbf{0.240} & \multicolumn{1}{c|}{\textbf{0.263}} & 0.258 & 0.278 & 0.272 & 0.291 & 0.258 & 0.280 & 0.259 & 0.286 & 0.309 & 0.360 \\
\multicolumn{1}{c|}{Exchange} & \underline0.350 & \multicolumn{1}{c|}{\underline0.396} & \textbf{0.338} & \multicolumn{1}{c|}{\textbf{0.390}} & 0.360 & 0.403 & 0.378 & 0.418 & 0.366 & 0.404 & 0.416 & 0.443 & 0.518 & 0.429 \\ \bottomrule
\end{tabular}
}
\end{table*}

\subsection{Imputation}\label{sec:exp:imputation}
Time series data from real-world systems often contain missing values, which can arise from sensor malfunctions or environmental conditions. These missing values complicate downstream analysis, necessitating imputation techniques in practical applications. For imputation to provide meaningful replacements for the missing data, the underlying architecture must effectively capture the temporal patterns inherent in the irregular and partially observed time series.

For the imputation task, a fixed portion $r_m \in [0,1]$ of values in the original multivariate input sequence is replaced by the masking value $-1$. The positions of replaced values are indicated by random imputation masks $\vm_1 \dots \vm_T$, where $\vm_i \in \{0,1\}^F$, $F$ is the number of features, and $r_m = \frac{1}{FT}\sum_{i=1}^{T}{\vm_i^\top \vm_i}$. This yields the multivariate masked
input sequence $\vx_1 \dots \vx_T$, where $\vx_i \in \mathbb{R}^F$. The prediction sequence $\hat{\vy}_{1} \dots \hat{\vy}_{H}$ here has the same length as the input sequence, thus $H=T$, and is aimed to accurately reconstruct the original multivariate sequence, that is, the ground truth $\vy_{1} \dots \vy_{T}$.

To measure the discrepancy between the prediction sequence $\hat{\vy}_{1} \dots \hat{\vy}_{T}$ and the ground truth $\vy_{1} \dots \vy_{
T}$, where $\hat{\vy}_{i}, \vy_{i} \in \mathbb{R}^F$, we chose the sum of the mean average error (MAE) and mean squared error (MSE) as the imputation loss $\mathcal{L}_\text{impt}$ during training. 
For the masked and unmasked regions separately, we calculate the losses $\mathcal{L}_\text{m}$ and $\mathcal{L}_\text{u}$, respectively. Both are averaged across all $F$ channels and $T$ timesteps. We adjust the contribution of the loss for the masked region by the weighting factor $\frac{1}{r_m}$. Consequently, we adapt its impact according to the missing ratio. In the following, $\mathbf{1}_F$ denotes the 1-vector of length $F$, that is, $\mathbf{1}_F =(1,...,1)^{\top}\in\mathbb{R}^F$:

\begin{equation}
\label{eq:imputation_loss}
\begin{aligned}
    \mathcal{L}_\text{m} &= 
    \frac{1}{r_m F T}
    \sum^{T}_{i=1}{
    ||\vm_i \odot (\hat{\vy}_{i}-\vy_{i})||_{1} +
    ||\vm_i \odot (\hat{\vy}_{i}-\vy_{i})||^{2}_{2}
    } \\
    \mathcal{L}_\text{u} &= 
    \frac{1}{(1-r_m) F T}
    \sum^{T}_{i=1}{
    ||(\mathbf{1}_F - \vm_i) \odot (\hat{\vy}_{i}-\vy_{i})||_{1} +
    ||(\mathbf{1}_F - \vm_i) \odot (\hat{\vy}_{i}-\vy_{i})||^{2}_{2}
    } \\
    \mathcal{L}_\text{imputation} &=
    \frac{1}{r_m} \mathcal{L}_\text{m} + \mathcal{L}_\text{u}
\end{aligned}
\end{equation}

To evaluate the performance of our architecture for TS imputation, we adopted the experimental setup by~\cite{wu2022timesnet} and introduced random data omissions into all datasets, resulting in four distinct missing rates: $r_m \in \{12.5\%, 25\%, 37.5\%, 50\% \}$.

For the evaluation, we consider six datasets (i.e., ECL, ETTm1, ETTm2, ETTh1, ETTh2, and Weather) and compare against multiple SOTA imputation techniques: TimesNet~\citep{wu2022timesnet}, LightTS~\citep{zhang2022less}, DLinear~\citep{zeng2022transformers}, and Stationary~\citep{liu2022non}.

\paragraph{Results}
We present the results of our proposed architecture in Table~\ref{tab:imputation1-avg}. On the ECL and Weather dataset, our architecture performs considerably well, whereas on the ETT datasets we were not able to match current SOTA results. Despite the good results with 26.03\% performance increase on the MSE metric for ECL compared to TimesNet, we report comparable decreases in performance on the ETT datasets. Despite this, our TSRM approach outperforms LightTS, and DLinear across all datasets for both averaged metrics. Similar to the forecasting experiments' results, the Weather dataset performs better with TSRM\_IFC than with TSRM, which reinforces the suspicion that inter-feature correlation plays a major role in this dataset.
We provide further details, including a separation into the different missing ratios and a comparison against additional SOTA approaches, in Appendix~\ref{app:imputation:results} and Table~\ref{app:tab:imputation1}. All reported results in Tables~\ref{tab:imputation1-avg}~and~\ref{app:tab:imputation1} were taken from~\cite{wu2022timesnet}.

\begin{table*}[]
\caption{Performance comparison for the multivariate imputation task with missing ratios $r_m\in\{0.125, 0.25, 0.375, 0.5\}$ and a fixed input length of 96. Results are averaged over all missing ratios. Bold/underline indicate best/second.}
\label{tab:imputation1-avg}
\centering
\scalebox{.88}{
\begin{tabular}{@{}ccccccccccccc@{}}
\toprule
Models & \multicolumn{2}{c}{TSRM} & \multicolumn{2}{c}{TSRM\_IFC} & \multicolumn{2}{c}{TimesNet} & \multicolumn{2}{c}{LigthTS} & \multicolumn{2}{c}{Dlinear} & \multicolumn{2}{c}{Stationary} \\ \midrule
\multicolumn{1}{c|}{Metrics} & MSE & \multicolumn{1}{c|}{MAE} & MSE & \multicolumn{1}{c|}{MAE} & MSE & MAE & MSE & MAE & MSE & MAE & MSE & MAE \\ \cmidrule(l){2-13} 
\multicolumn{1}{c|}{ECL} & \textbf{0.073} & \multicolumn{1}{c|}{\textbf{0.170}} & \textbf{0.073} & \multicolumn{1}{c|}{\underline0.179} & 0.092 & 0.210 & 0.131 & 0.262 & 0.132 & 0.260 & 0.100 & 0.218 \\
\multicolumn{1}{c|}{ETTm1} & 0.042 & \multicolumn{1}{c|}{\underline0.116} & 0.045 & \multicolumn{1}{c|}{0.140} & \textbf{0.027} & \textbf{0.107} & 0.104 & 0.218 & 0.093 & 0.206 & \underline0.036 & 0.126 \\
\multicolumn{1}{c|}{ETTm2} & 0.028 & \multicolumn{1}{c|}{0.103} & \underline0.024 & \multicolumn{1}{c|}{0.100} & \textbf{0.022} & \textbf{0.088} & 0.046 & 0.151 & 0.096 & 0.208 & 0.026 & \underline0.099 \\
\multicolumn{1}{c|}{ETTh1} & 0.102 & \multicolumn{1}{c|}{0.212} & \textbf{0.059} & \multicolumn{1}{c|}{\textbf{0.165}} & \underline0.078 & \underline0.187 & 0.284 & 0.374 & 0.201 & 0.306 & 0.094 & 0.202 \\
\multicolumn{1}{c|}{ETTh2} & 0.072 & \multicolumn{1}{c|}{0.173} & \underline0.051 & \multicolumn{1}{c|}{\underline0.151} & \textbf{0.050} & \textbf{0.146} & 0.120 & 0.250 & 0.142 & 0.260 & 0.053 & 0.152 \\
\multicolumn{1}{c|}{Weather} & 0.031 & \multicolumn{1}{c|}{\underline0.048} & \textbf{0.029} & \multicolumn{1}{c|}{\textbf{0.045}} & \underline0.030 & 0.054 & 0.056 & 0.116 & 0.052 & 0.110 & 0.032 & 0.059 \\ \bottomrule
\end{tabular}
}
\end{table*}

\subsection{Complexity Analysis}
Our proposed architecture exhibits a stable performance that meets or even exceeds SOTA results, especially for TS forecasting. However, in addition to the pure performance metrics, a model's complexity should also play an essential role when assessing an architecture's quality, not least in order to ensure its cost efficiency and applicability in practice. 
Our architecture is based on layers with comparably small number of trainable parameters. Furthermore, it is designed for memory efficiency having a small memory footprint. For example, a model for the ETTh1 dataset in the prediction task (96/96) with 8 batches requires only 500 MB of GPU memory in total. In Table~\ref{tab:exp:parameters}, we compare the number of trainable parameters of TSRM against SOTA architectures most similar to ours, as an indicator for complexity. The reported values for the SOTA approaches originate from the work of~\cite{wang2024tssurvey}, the Time-Series-Library\footnote{https://github.com/thuml/Time-Series-Library}, and were collected separately for all datasets and the corresponding configuration. All reported trainable parameters of TSRM were collected from the corresponding TS forecasting models reported in Table~\ref{tab:forecasting1-avg}. To ensure comparability, the lookback and prediction horizon were always set to 96. The last column in Table~\ref{tab:exp:parameters} shows the average number of parameters across datasets. While most models, on average, have more than 10 million trainable parameters (median $6.903\text{M}\pm27.648\text{M}$), TSRM exhibits a much lower complexity with only a few hundred thousand trainable parameters for all except the ETTm2 dataset. 

\begin{table*}[]
\centering
\caption{Amount of trainable parameters in million. Lookback window is fixed to 96.}
\label{tab:exp:parameters}
\scalebox{.85}{
\begin{tabular}{@{}ccccccccc@{}}
\toprule
Model &  ETTh1  & ETTm2  & ECL     & Exchange & Weather &  & Average   \\
\cmidrule(lr){1-1}
\cmidrule(lr){2-2}
\cmidrule(lr){3-3}
\cmidrule(lr){4-4}
\cmidrule(lr){5-5}
\cmidrule(lr){6-6}
\cmidrule(lr){8-8}
TimesNet     & \underline{0.605}  & \textbf{1.191}  & 150.304 & 4.708    & 1.193   && 31.600 \\
Autoformer   & 10.535 & 10.535 & 12.143  & 10.541   & 10.607  && 10.872 \\
Transformer  & 10.540 & 10.540 & 10.518  & 10.543   & 10.590  && 10.546 \\
PatchTST     & 3.751  & 10.056 & 6.903   & 6.903    & 6.903   && 6.903  \\
Crossformer  & 42.176 & 42.139 & 9.261   & 0.437    & \textbf{0.123}   && 18.827 \\
iTransformer & \textbf{0.224}  & 4.833  & \underline{4.833}   & \textbf{0.224}    & 4.833   && \underline{2.989}  \\
TSRM         & 0.857  & \underline{2.781}  & \textbf{0.161}   & \underline{0.382}    & \underline{0.338}   && \textbf{0.904}  \\ \bottomrule
\end{tabular}
}
\end{table*}
\section{Ablation Study}\label{sec:ablation}

\begin{figure}
\centering
\begin{subfigure}{}
  \centering
    \includegraphics[width=.49\linewidth]{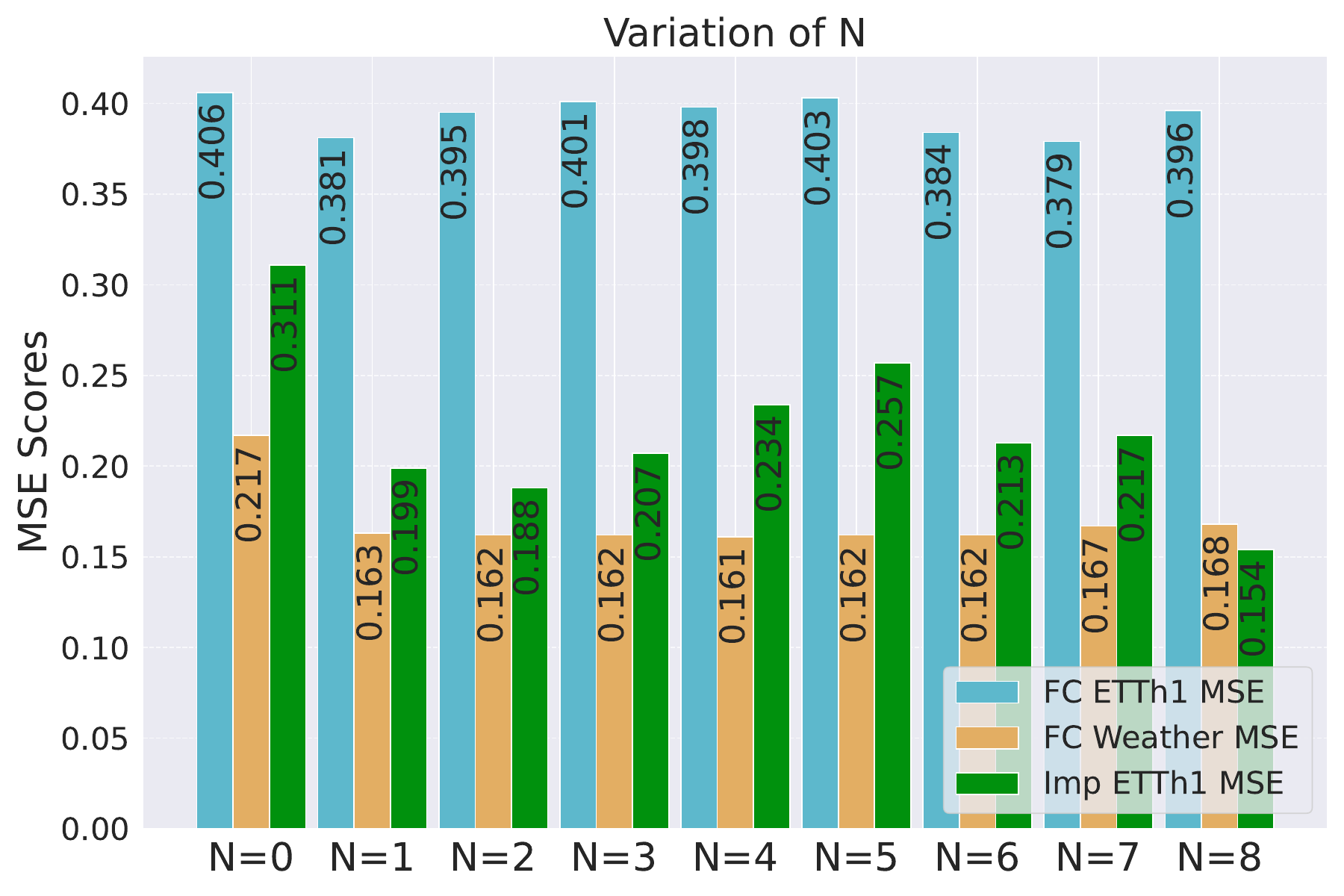}
\end{subfigure}%
\begin{subfigure}{}
  \centering
    \includegraphics[width=.49\linewidth]{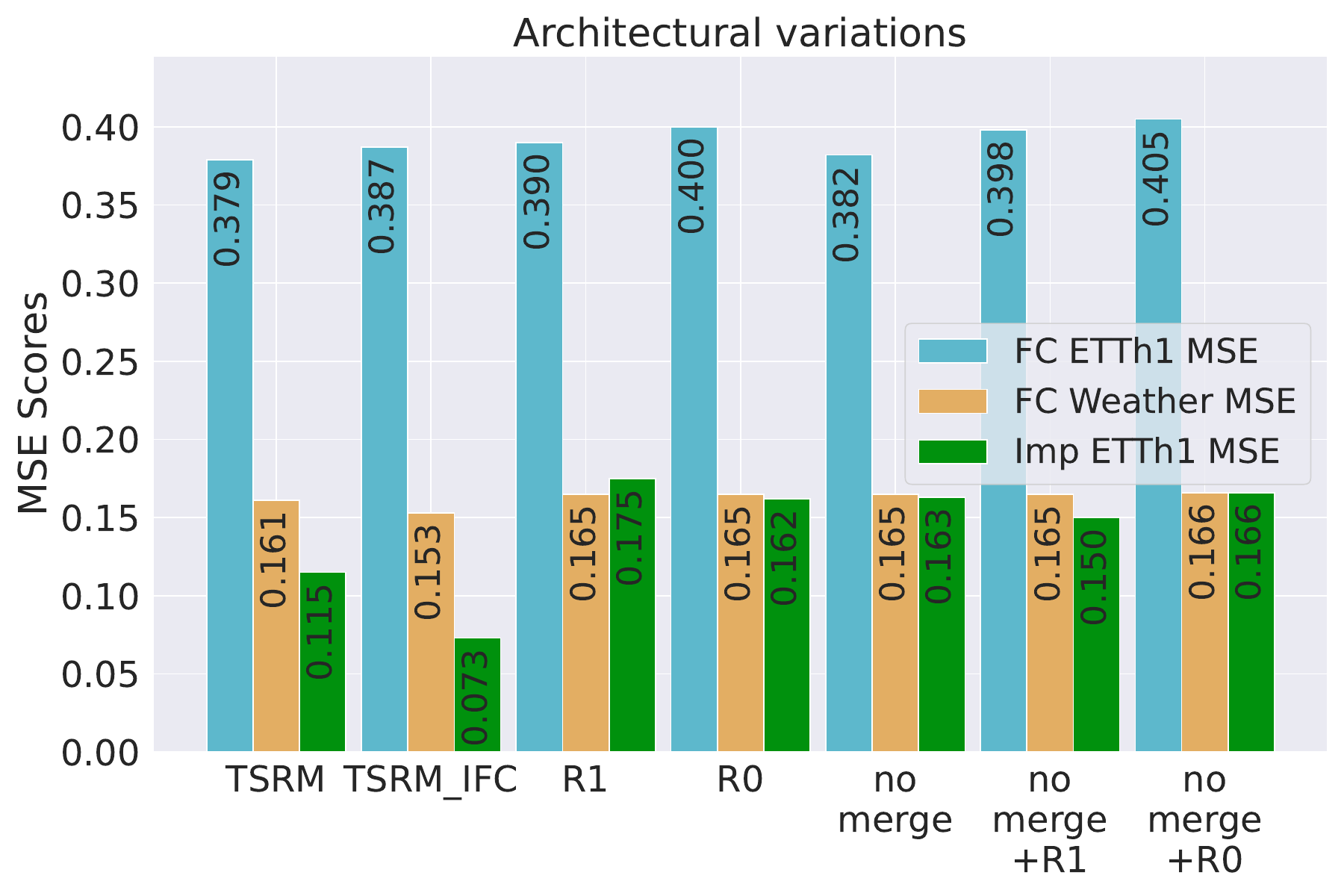}

\end{subfigure}
\caption{Ablation study results for the architecture variations (bottom) and sensitivity study for the hyperparameter $N$ (top) on the datasets Weather and ETTh1.}
\label{fig:ablation}

\end{figure}

To evaluate the contribution of different modules within TSRM, we conduct ablation studies centered around the core component \EncodingLayer{}. In particular, we investigate the effects of varying the number of stacked \EncodingLayer{}s through a sensitivity analysis of $N$ and change its components. For all experiments, we use ETTh1 as the dataset, which performs well without feature interaction (TSRM), and Weather, which shows better results when used with TSRM\_IFC, that is, with feature interaction. To gain further insights into the architecture, we carry out the experiments with forecasting and imputation.

\paragraph{Varying the number of \EncodingLayer{}s.} We investigate the sensitivity of our architecture towards the amount of \EncodingLayer{}s ($N$). Therefore, we picked the best hyperparamter constellations of the forecasting task and run experiments with $N \in \{0,1,2,\dots 8\}$. Figure~\ref{fig:ablation} (top) shows the results for all variations. The configuration of $N=0$ means that no \EncodingLayer{} is involved. For both datasets on the forecasting task (FC), we observe a plateau of the MSE metric in certain ranges. For example, the ETTh1 dataset performs best with a fairly high number of $N$, while Weather performs best with $N=4$ and then gets worse as $N$ increases. For both datasets, however, the results are significantly better when using at least one \EncodingLayer{}. It becomes clearer with the imputation task. It shows the clear superiority of several stacked \EncodingLayer{} and a steady improvement in performance along the x-axis. 
The configuration without any \EncodingLayer{} ($N=0$) yields the worst result on the ETTh1 dataset. However, it is worth noting that despite the lack of any complex logic in the case of $N=0$, the result is comparable to DLinear on the weather dataset and on the ETTh1 dataset. This can be explained by the fact that the remaining architecture is comparable to that of DLinear. More details can be found in Appendix~\ref{app:ablation}.

\paragraph{Architecture variations} We explore the influence of various architectural components on the performance of TSRM. Specifically, we assess the effect of the \MergeLayer{} by blocking gradient flow to prevent the layer from learning aggregations in the experiment \textit{no\_merge}. Like this, the \MergeLayer{} can still fulfill its role of restoring the original dimensions of the sequence, allowing the \EncodingLayer{}s to remain stackable, albeit without trainable parameters. Additionally, we reduce $R$ to a single CNN layer with a kernel size of three and a dilation of one in the experiment \textit{R1}, and with a deactivated \MergeLayer{} in the experiment \textit{no\_merge+R1}. In a further step, we restrict the CNN layer to a kernel size of one, eliminating any structural learning effects and limiting its function to position-wise weighting, as tested in the experiment \textit{R0}, and with a deactivated \MergeLayer{} in the experiment \textit{no\_merge+R0}. 

Figure~\ref{fig:ablation} (bottom) shows the results of all ablation experiments. 
A decline in performance of the TSRM architecture is evident throughout the ablation experiments. The most notable drop occurs between the original TSRM architecture and the \textit{R0} experiment, where a considerable loss in performance highlights the critical role of the CNN layers within \RepresentationLayer{}. As with the previous experiment, the imputation shows a clear change in performance and demonstrates the importance of the individual components.

The performance changes without the trainable \MergeLayer{} are not significant compared to the original TSRM architecture. However, it is important to note that the role of \MergeLayer{} is mainly to undo the dimensional changes caused by the representation layer, a prerequisite for stacking several encoding layers, regardless of the input length. Reducing the number of 1D convolutions (i.e., $R=1$) and preventing structural learning with a kernel size of 1, further reduces performance, although the drop is less pronounced than between the scenarios with and without \MergeLayer{}s. 

Interestingly, the experiment \textit{no\_merge+R0}, in which the architectural structure is changed, is very similar to that of iTransformer~\citep{liu2024itransformer} and leads to comparable performance results, as shown in Appendix~\ref{app:fc:results} (Table~\ref{app:tab:forecasting1}). The weather dataset shows no performance changes between the experiments \textit{no\_merge} and \textit{no\_merge+$R1$}, which can be explained by the fact that the most powerful model in this setup and thus, the configuration used in this experiment, uses the same CNN configuration as in $R1$, which accordingly leads to no performance changes. More details can be found in Appendix~\ref{app:ablation}.

\section{Conclusion}
We introduced a new architecture for time series forecasting and imputation, the Time Series Representation Model (TSRM). This model uses hierarchically organized encoding layers (\EncodingLayer{}) designed to independently learn representations from the input sequence at different levels of abstraction, with each layer passing learned and aggregated features to the next. \EncodingLayer{} is largely based on the concept of self-attention and consists of a representation layer and an aggregation layer, which are responsible for representing the input sequence at different levels of abstraction as well as aggregating and embedding the learned or highlighted representations.
The architecture is designed to be of low complexity while supporting explainability in the form of detailed attention highlighting.
Our empirical evaluation showed that TSRM is able to outperform SOTA approaches on a number of well-established benchmark datasets in the area of time series forecasting and imputation. Moreover, it significantly reduces complexity with respect to the number of trainable parameters.
In future work, we plan to evaluate the architecture regarding a pretraining/fine-tuning, few/zero-shot learning, and foundation model approach, as well as for further tasks such as classification and anomaly detection. 
%


\section*{Impact Statement}
In this work, we contribute to the advancement of machine learning by introducing a novel architecture for time series forecasting and imputation, demonstrating that lightweight self-attention-based models can achieve competitive performance while maintaining computational efficiency. Additionally, the architecture is explicitly designed for interpretability, providing deeper insights into the model’s training dynamics. This enables systematic evaluation and optimization, improving adaptability and accelerating the model selection and refinement process.
There are many potential societal impacts of our work that we believe do not need to be highlighted here.

\bibliographystyle{unsrtnat}


\newpage
\appendix

\section{Related Work}
\label{sec:rw}

\textbf{Transformer-based models.} Since its inception in 2017, Transformer~\citep{vaswani2017attention} and its numerous derivatives~\citep{zhou2021informer, wu2021autoformer, zhou2022fedformer, nie2022time, du2023saits, zhang2023crossformer, chen2024pathformer, zhao2024himtm, das2024decoder, liu2024itransformer} steadily gained traction and are now a well-established approach to time series modeling. One of the more recent works is PatchTST~\citep{nie2022time}, which combines the Transformer encoder with subseries-level patches as input encoding to increase efficiency while demonstrating strong modeling capacity. While PatchTST processes each channel of multivariate TS independently, Crossformer~\citep{zhang2023crossformer} captures both temporal and cross-channel dependencies. To this end, the model unravels the input TS into two dimensions and features a novel attention layer to learn both types of dependencies efficiently. Pathformer~\citep{chen2024pathformer} is a multiscale transformer with adaptive dual attention to capture temporal dependencies between TS segments of varying granularity. In contrast, our proposed TSRM, while using classical multi-head self-attention internally, does not utilize the Transformer architecture but an adaptation of the Transformer encoder only, to limit the memory footprint and reduce complexity.

\textbf{Self-supervised pretraining.} Splitting the training process into pretraining and fine-tuning allows TS models to learn universal representations that can be later utilized for different downstream tasks~\citep{jiang2022transferability}. One work in this field was proposed by~\cite{ekambaram2023tsmixer}. Their TSMixer is built around an MLP backbone, while our approach employs a convolution- and self-attention-based encoder architecture, such as SimMTM~\citep{dong2023simmtm} and its successor HiMTM~\citep{zhao2024himtm}. In contrast, CoST~\citep{woo2022cost} learns disentangled feature representations by discriminating the trend and seasonal components. \cite{lee2024learning} recently presented PITS. 

\textbf{Foundation models.} Similar to self-supervised pretraining, time series foundation models learn universal representations of TS and use them for different downstream tasks~\citep{bommasani2021opportunities}. However, they are more powerful in that they pretrain on a cross-domain database to generalize across individual target datasets. In recent years, various approaches have been proposed, including TF-C~\citep{zhang2022self}, TimesNet~\citep{wu2022timesnet}, FPT~\citep{zhou2023one}, Lag-Llama~\citep{rasul2023lag}, MOMENT~\citep{goswami2024moment}, MOIRAI~\citep{woo2024unified}, and TimesFM~\citep{das2024decoder}. TimesFM and FPT are Transformer-based models. TF-C employs a different embedding stage than TSRM, which is based on time-frequency-consistency and contrastive learning. Where TimesNet analyses temporal variations in the 1D input sequence by unfolding it into two dimensions along multiple periods observed over the time axis, TSRM is designed as multilayered representation architecture and it embeds temporal variations into a one-dimensional vector (separately for each layer). MOIRAI follows a patch-based approach with a masked encoder architecture. Compared to TSRM, MOMENT differs by using patching, a Transformer encoder directly, and self-supervised pretraining on a wide range of datasets.

\textbf{Patch-based models.} Patching is a form of input encoding that divides the time series into subsequences, which can be either overlapping or non-overlapping~\citep{nie2022time, zhang2023crossformer, ekambaram2023tsmixer, zhou2023one, das2024decoder, lee2024learning, chen2024pathformer, liu2024itransformer, goswami2024moment, woo2024unified}. In the basic form, identical-sized patches are sliced from the input TS and fed as tokens to the model~\citep{nie2022time}. Pathformer's multiscale division divides the TS into different temporal resolutions using patches of various, dynamically chosen sizes. Crossformer computes more complex patches, encoding both temporal and cross-channel dependencies. iTransformer~\citep{liu2024itransformer} takes the idea to the extreme, operating on patches covering an entire channel of the input TS each. What sets TSRM apart from previous works is the representation learning with multilayered and multidimensional CNN layers, dynamically learned in a novel representation layer, in order to cover different granularities and enable hierarchical representation learning.

\textbf{Few/zero-shot learning.} Few-shot learning refers to the capability of a model to generalize from the data domain it is (pre-)trained on to a new target domain using just a few (zero-shot learning: none) target-training instances~\citep{zhou2023one, rasul2023lag, das2024decoder, lee2024learning, woo2024unified}. Lag-Llama is based on a decoder-only Transformer architecture that uses lags as covariates and processes only univariate TS, while TSRM uses CNN-extracted feature vectors and can handle multivariate TS. Moreover, Lag-Llama is pretrained on a large corpus of multidomain TS data, while FPT utilizes a pretrained language model like BERT~\citep{devlin2018bert} as basis. In this work, however, we do not consider the few/zero-shot setting, but instead, we train and optimize TSRM on each dataset separately, focusing on deriving profound representation models that incorporate knowledge about each specific type of TS.


\textbf{Delimitation.} Our presented TSRM architecture differs from the above-mentioned work by combining novel approaches, such as the temporal feature extraction of different abstractions like in TimesNet\citep{wu2022timesnet} and the self-attention approach in Crossformer~\citep{zhang2023crossformer} or PatchTST~\citep{nie2022time}. Furthermore, our architecture design allows an independent stacking of the individual representation learning layers to allow hierarchical learning and a significant reduction of complexity with respect to the learnable parameters.



\section{Datasets}\label{appendix:datasets}

Below, we provide more details on the datasets used in our experiments. Please find a detailed overview of all employed benchmark datasets in Table~\ref{tab:datasets}.

\textbf{Electricity Load Diagram (ECL)}: 
The Electricity dataset, available at UCI~\citep{dua2017uci}, contains electricity consumption data measured in kilowatt-hours (kWh). It includes data from 370 clients collected every 15 minutes for 48 months, starting from January 2011 to December 2014. 

\textbf{Weather}: The weather dataset contains the recordings of 21 meteorological factors, such as temperature, humidity, and air pressure, collected every 10 minutes from the weather station of the Max Planck Biogeochemistry Institute in Jena, Germany in 2020~\citep{wu2021autoformer}.

\textbf{Exchange}: This dataset collects the daily exchange rates of 8 different currencies (Australia, British, Canada, Switzerland, China, Japan, New Zealand, and Singapore) from 1990 to 2016~\citep{wu2021autoformer}.

\textbf{Electricity Transformer Temperature (ETT)}:
The ETT dataset comprises data collected from electricity transformers over a time period from July 1, 2016, to June 26, 2018. ETT consists of 4 subsets, where ETTh1 and ETTh2 contain records with hourly resolution, while ETTm1 and ETTm2 are recorded every 15 minutes. In total, ETT includes 69,680 data points without any missing values. Each record contains seven features, including oil temperature and six different types of external power load features~\citep{zhou2021informer}. 

\begin{table}[ht]
\caption{Details of the used benchmark datasets. The assignment to train, validation, or test follows the established procedure~\citep{wu2021autoformer}.}
\label{tab:datasets}
\centering
\scalebox{.85}{
\begin{tabular}{@{}lllll@{}}
\toprule
Dataset & Channels & Size (train / val / test) & Frequency & Information \\ \midrule
ECL & 321 & 18317 / 2633 / 5261 & Hourly & Electricity \\
ETTm1,ETTm2 & 7 & 34465 / 11521 / 11521 & 15min & Electricity \\
ETTh1,ETTh2 & 7 & 8545 / 2881 / 2881 & Hourly & Electricity \\
Exchange & 8 & 5120 / 665 / 1422 & Daily & Economy \\
Weather & 21 & 36792 / 5271 / 10540 & 10min & Weather \\
\bottomrule
\end{tabular}
}
\end{table}

\section{Explainability with Attention Weight Highlighting}\label{appendix:exp_explain1}
Our methodology's fundamental architectural principle is predicated on utilizing the attention mechanism as its central component and maintaining dimensional consistency throughout all \EncodingLayer{}s. 
As detailed in Section~\ref{sec:methodology}, the attention layers play a pivotal role in enhancing the representations from the representation layers.
This approach, combined with the low complexity of our architecture, enables us to extract and investigate the attention weights of all \EncodingLayer{}s, offering valuable insights into our architecture's functioning and decision making. 
Due to its design, we are able to extract and analyze individual attention weights for all $N$ \EncodingLayer{}s' attention layers and all $F$ features individually. In other words, separate attention weights can be generated for each feature and \EncodingLayer{}, thereby enabling their analysis in isolation and combined. 
For this, 
it is imperative to revert the matrix dimensions dictated by the \RepresentationLayer{}s back to those of the input TS. This transformation employs the identical transpose CNN layer utilized in the \MergeLayer{}s, albeit with static weight matrices, designed to calculate the mean attention weight for each value. 
The $N$ back-transformed attention matrices can then be visualized together with the output sequence to analyze the architecture's weighting during imputation, or forecasting. This can be done for all $N$ \EncodingLayer{}s and features individually or as a sum over all \EncodingLayer{}s to get an overview of all weights. 

\begin{figure}
     \centering
     \begin{subfigure}
         \centering
         $N=0$
         \includegraphics[width=\textwidth]{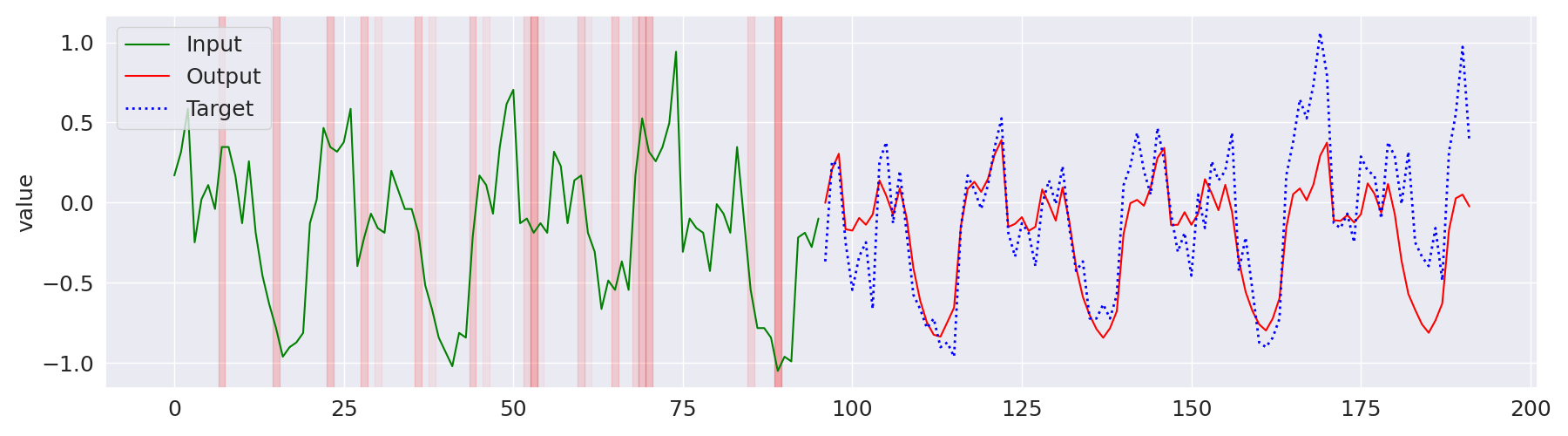}
        
     \end{subfigure} 
     \vfill 
     \begin{subfigure}
         \centering
         $N=1$
         \includegraphics[width=\textwidth]{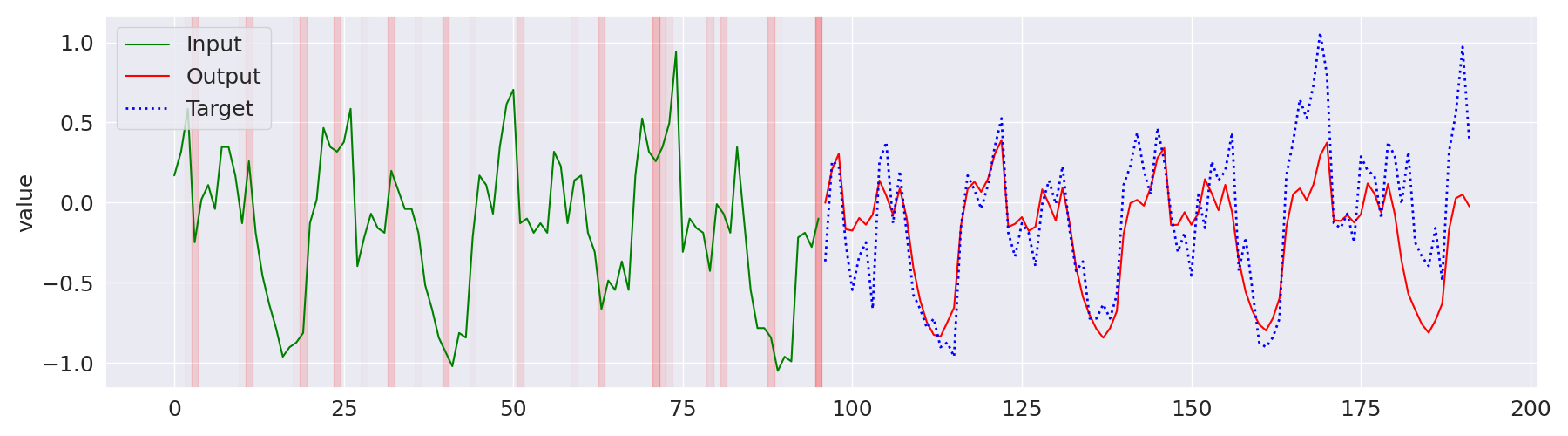}
     \end{subfigure}
     \vfill 
     \begin{subfigure}
         \centering
        $N=2$
         \includegraphics[width=\textwidth]{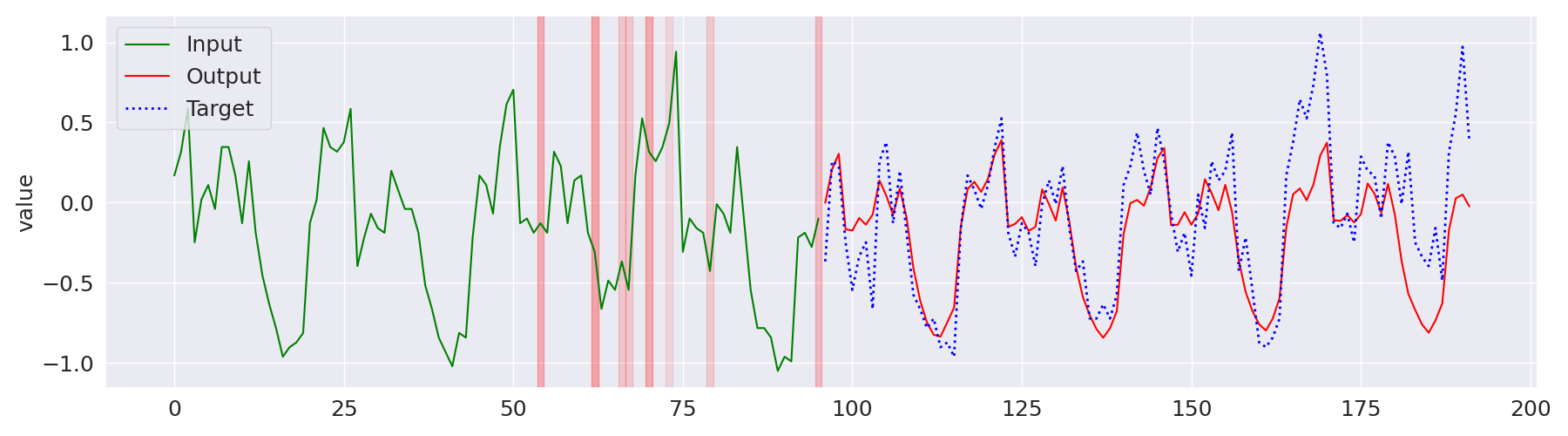}
     \end{subfigure}
     \vfill 
     \begin{subfigure}
         \centering
         Summed and normalized combination over all $N$
         \includegraphics[width=\textwidth]{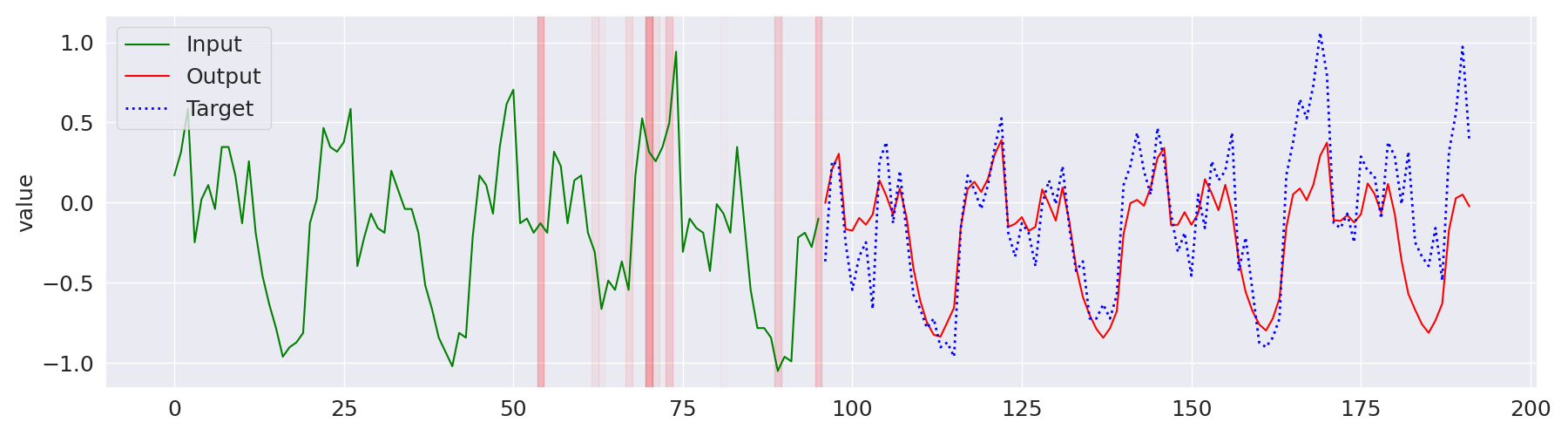}
     \end{subfigure}
        \caption{Highlighted attention weights during an ETTh1 forecasting task for all 3 \EncodingLayer{}s, starting with the first \EncodingLayer{} at the top and concluding with the the combined version over all \EncodingLayer{} at the bottom. }
\label{fig:appendix:explain}
\end{figure}

Figure~\ref{fig:appendix:explain} shows an example using the ETTh1 dataset on the first feature during a forecasting task with all three \EncodingLayer{}s separately visualized, including the combined attention weights at the bottom. The solid green line represents the initial input series, followed by a dotted blue line after the 96th value. This dotted blue trajectory delineates the target horizon. The red line indicates the prediction of the model. The emphasis of attention is subject to a threshold value of 0.85 (normalized) in order to emphasize only the most important aspects of attention.
We can observe that the attention from the first \EncodingLayer{} is more distributed and mainly focuses on high and low points in the sequence. 
Later, attention seems to be more dense and switches on selected representatives of repeating sub-areas in the time series, e.g. the pattern around 75, as well as on the last known value before the horizon, giving the impression that it is focusing on these areas more closely. 
However, it should be noted that this is not an evaluation but rather an interpretation of a snapshot, and cannot be taken as evidence of true explainability.

\section{Forecasting}\label{app:fc:results}

We extend our evaluation from Section~\ref{sec:exp:fc} with a more detailed evaluation, including more SOTA approaches for the comparison, and we provide the results of all prediction lengths separately.
For the evaluation, we consider seven datasets (i.e., ECL, ETTm1, ETTm2, ETTh1, ETTh2, Weather, and Exchange) and compare against multiple SOTA techniques: iTransformer~\citep{liu2024itransformer}, RLinear~\citep{li2023revisiting}, PatchTST~\citep{nie2022time}, Crossformer~\citep{zhang2023crossformer}, TiDE~\citep{das2023long}, TimesNet~\citep{wu2022timesnet}, DLinear~\citep{zeng2023transformers}, SCINet~\citep{liu2022scinet}, FEDformer~\citep{zhou2022fedformer}, Stationary~\citep{liu2022non}, and Autoformer~\citep{wu2021autoformer}.

Table~\ref{app:tab:forecasting1} shows the extended results with all four prediction horizons ($H$) separated. We further report the training and inference time for the ETTh1 dataset with the best configuration and a horizon of 96 with 28 seconds per epoch for training and 6 seconds per epoch for inference. For the weather dataset, also with the best configuration and a horizon of 96, we report 194 seconds per epoch during training and 52 seconds per epoch during inference. 

\begin{table}
\centering
\caption{Performance comparison for the multivariate forecasting task with prediction horizons $H \in \{96, 192, 336, 720\}$ and fixed lookback window $T = 96$. AVG shows the averaged result over all prediction horizons per dataset and model. Bold/underline indicate best/second.}
\label{app:tab:forecasting1}
\scalebox{.49}{
\begin{tabular}{@{}cccccccccccccccccccccccccccc@{}}
\toprule
\multicolumn{2}{c}{Models} & \multicolumn{2}{c}{TSRM} & \multicolumn{2}{c}{TSRM\_M} & \multicolumn{2}{c}{iTransformer} & \multicolumn{2}{c}{RLinear} & \multicolumn{2}{c}{PatchTST} & \multicolumn{2}{c}{Crossformer} & \multicolumn{2}{c}{TiDE} & \multicolumn{2}{c}{TimesNet} & \multicolumn{2}{c}{DLinear} & \multicolumn{2}{c}{SCINet} & \multicolumn{2}{c}{FEDformer} & \multicolumn{2}{c}{Stationary} & \multicolumn{2}{c}{Autoformer} \\ \midrule
\multicolumn{2}{c|}{Metrics} & MSE & \multicolumn{1}{c|}{MAE} & MSE & \multicolumn{1}{c|}{MAE} & MSE & MAE & MSE & MAE & MSE & MAE & MSE & MAE & MSE & MAE & MSE & MAE & MSE & MAE & MSE & MAE & MSE & MAE & MSE & MAE & MSE & MAE \\
\multirow{5}{*}{\rotatebox{90}{\small{ECL}}} & \multicolumn{1}{c|}{96} & 0.168 & \multicolumn{1}{c|}{0.255} & \textbf{0.148} & \multicolumn{1}{c|}{\underline{0.245}} & \textbf{0.148} & \textbf{0.240} & 0.201 & 0.281 & 0.181 & 0.270 & 0.219 & 0.314 & 0.237 & 0.329 & 0.168 & 0.272 & 0.197 & 0.282 & 0.247 & 0.345 & 0.193 & 0.308 & 0.169 & 0.273 & 0.201 & 0.317 \\
 & \multicolumn{1}{c|}{192} & \underline{0.176} & \multicolumn{1}{c|}{\underline{0.262}} & \underline{0.176} & \multicolumn{1}{c|}{\underline{0.262}} & \textbf{0.162} & \textbf{0.253} & 0.201 & 0.283 & 0.188 & 0.274 & 0.231 & 0.322 & 0.236 & 0.330 & 0.184 & 0.289 & 0.196 & 0.285 & 0.257 & 0.355 & 0.201 & 0.315 & 0.182 & 0.286 & 0.222 & 0.334 \\
 & \multicolumn{1}{c|}{336} & 0.192 & \multicolumn{1}{c|}{0.278} & \textbf{0.175} & \multicolumn{1}{c|}{\underline{0.274}} & \underline{0.178} & \textbf{0.269} & 0.215 & 0.298 & 0.204 & 0.293 & 0.246 & 0.337 & 0.249 & 0.344 & 0.198 & 0.300 & 0.209 & 0.301 & 0.269 & 0.369 & 0.214 & 0.329 & 0.200 & 0.304 & 0.231 & 0.338 \\
 & \multicolumn{1}{c|}{720} & 0.234 & \multicolumn{1}{c|}{\underline{0.312}} & \textbf{0.202} & \multicolumn{1}{c|}{\textbf{0.299}} & 0.225 & 0.317 & 0.257 & 0.331 & 0.246 & 0.324 & 0.280 & 0.363 & 0.284 & 0.373 & \underline{0.220} & 0.320 & 0.245 & 0.333 & 0.299 & 0.390 & 0.246 & 0.355 & 0.222 & 0.321 & 0.254 & 0.361 \\
 & \multicolumn{1}{c|}{AVG} & 0.192 & \multicolumn{1}{c|}{0.277} & \textbf{0.175} & \multicolumn{1}{c|}{\textbf{0.270}} & \underline{0.178} & \textbf{0.270} & 0.218 & 0.298 & 0.205 & 0.290 & 0.244 & 0.334 & 0.252 & 0.344 & 0.192 & 0.295 & 0.212 & 0.300 & 0.268 & 0.365 & 0.214 & 0.327 & 0.193 & 0.296 & 0.227 & 0.338 \\
 & \multicolumn{1}{c|}{} &  & \multicolumn{1}{c|}{} &  & \multicolumn{1}{c|}{} &  &  &  &  &  &  &  &  &  &  &  &  &  &  &  &  &  &  &  &  &  &  \\
\multirow{5}{*}{\rotatebox{90}{\small{ETTm1}}} & \multicolumn{1}{c|}{96} & \textbf{0.314} & \multicolumn{1}{c|}{\textbf{0.352}} & \textbf{0.314} & \multicolumn{1}{c|}{\underline{0.354}} & 0.334 & 0.368 & 0.355 & 0.376 & 0.329 & 0.367 & 0.404 & 0.426 & 0.364 & 0.387 & 0.338 & 0.375 & 0.345 & 0.372 & 0.418 & 0.438 & 0.379 & 0.419 & 0.386 & 0.398 & 0.505 & 0.475 \\
 & \multicolumn{1}{c|}{192} & \underline{0.364} & \multicolumn{1}{c|}{\textbf{0.377}} & \textbf{0.363} & \multicolumn{1}{c|}{\underline{0.382}} & 0.377 & 0.391 & 0.391 & 0.392 & 0.367 & 0.385 & 0.450 & 0.451 & 0.398 & 0.404 & 0.374 & 0.387 & 0.380 & 0.389 & 0.439 & 0.450 & 0.426 & 0.441 & 0.459 & 0.444 & 0.553 & 0.496 \\
 & \multicolumn{1}{c|}{336} & \textbf{0.393} & \multicolumn{1}{c|}{\textbf{0.401}} & \underline{0.398} & \multicolumn{1}{c|}{\underline{0.403}} & 0.426 & 0.420 & 0.424 & 0.415 & 0.399 & 0.410 & 0.532 & 0.515 & 0.428 & 0.425 & 0.410 & 0.411 & 0.413 & 0.413 & 0.490 & 0.485 & 0.445 & 0.459 & 0.495 & 0.464 & 0.621 & 0.537 \\
 & \multicolumn{1}{c|}{720} & \textbf{0.448} & \multicolumn{1}{c|}{\textbf{0.435}} & 0.461 & \multicolumn{1}{c|}{\underline{0.437}} & 0.491 & 0.459 & 0.487 & 0.450 & \underline{0.454} & 0.439 & 0.666 & 0.589 & 0.487 & 0.461 & 0.478 & 0.450 & 0.474 & 0.453 & 0.595 & 0.550 & 0.543 & 0.490 & 0.585 & 0.516 & 0.671 & 0.561 \\
 & \multicolumn{1}{c|}{AVG} & \textbf{0.380} & \multicolumn{1}{c|}{\textbf{0.391}} & \underline{0.384} & \multicolumn{1}{c|}{\underline{0.394}} & 0.407 & 0.410 & 0.414 & 0.408 & 0.387 & 0.400 & 0.513 & 0.495 & 0.419 & 0.419 & 0.400 & 0.406 & 0.403 & 0.407 & 0.486 & 0.481 & 0.448 & 0.452 & 0.481 & 0.456 & 0.588 & 0.517 \\
 & \multicolumn{1}{c|}{} &  & \multicolumn{1}{c|}{} &  & \multicolumn{1}{c|}{} &  &  &  &  &  &  &  &  &  &  &  &  &  &  &  &  &  &  &  &  &  &  \\
\multirow{5}{*}{\rotatebox{90}{\small{ETTm2}}} & \multicolumn{1}{c|}{96} & \underline{0.172} & \multicolumn{1}{c|}{\underline{0.256}} & \textbf{0.169} & \multicolumn{1}{c|}{\textbf{0.253}} & 0.180 & 0.264 & 0.182 & 0.265 & 0.175 & 0.259 & 0.287 & 0.366 & 0.207 & 0.305 & 0.187 & 0.267 & 0.193 & 0.292 & 0.286 & 0.377 & 0.203 & 0.287 & 0.192 & 0.274 & 0.255 & 0.339 \\
 & \multicolumn{1}{c|}{192} & \underline{0.238} & \multicolumn{1}{c|}{\underline{0.300}} & \textbf{0.236} & \multicolumn{1}{c|}{\textbf{0.297}} & 0.250 & 0.309 & 0.246 & 0.304 & 0.241 & 0.302 & 0.414 & 0.492 & 0.290 & 0.364 & 0.249 & 0.309 & 0.284 & 0.362 & 0.399 & 0.445 & 0.269 & 0.328 & 0.280 & 0.339 & 0.281 & 0.340 \\
 & \multicolumn{1}{c|}{336} & \underline{0.304} & \multicolumn{1}{c|}{\underline{0.341}} & \textbf{0.296} & \multicolumn{1}{c|}{\textbf{0.338}} & 0.311 & 0.348 & 0.307 & 0.342 & 0.305 & 0.343 & 0.597 & 0.542 & 0.377 & 0.422 & 0.321 & 0.351 & 0.369 & 0.427 & 0.637 & 0.591 & 0.325 & 0.366 & 0.334 & 0.361 & 0.339 & 0.372 \\
 & \multicolumn{1}{c|}{720} & \textbf{0.398} & \multicolumn{1}{c|}{\textbf{0.397}} & 0.404 & \multicolumn{1}{c|}{\textbf{0.397}} & 0.412 & 0.407 & 0.407 & 0.398 & \underline{0.402} & 0.400 & 1.730 & 1.042 & 0.558 & 0.524 & 0.408 & 0.403 & 0.554 & 0.522 & 0.960 & 0.735 & 0.421 & 0.415 & 0.417 & 0.413 & 0.433 & 0.432 \\
 & \multicolumn{1}{c|}{AVG} & \underline{0.278} & \multicolumn{1}{c|}{\underline{0.324}} & \textbf{0.276} & \multicolumn{1}{c|}{\textbf{0.321}} & 0.288 & 0.332 & 0.286 & 0.327 & 0.281 & 0.326 & 0.757 & 0.610 & 0.358 & 0.404 & 0.291 & 0.332 & 0.350 & 0.401 & 0.570 & 0.537 & 0.304 & 0.349 & 0.306 & 0.347 & 0.327 & 0.371 \\
 & \multicolumn{1}{c|}{} &  & \multicolumn{1}{c|}{} &  & \multicolumn{1}{c|}{} &  &  &  &  &  &  &  &  &  &  &  &  &  &  &  &  &  &  &  &  &  &  \\
\multirow{5}{*}{\rotatebox{90}{\small{ETTh1}}} & \multicolumn{1}{c|}{96} & \underline{0.377} & \multicolumn{1}{c|}{\underline{0.396}} & 0.379 & \multicolumn{1}{c|}{0.403} & 0.386 & 0.405 & 0.386 & \textbf{0.395} & 0.414 & 0.419 & 0.423 & 0.448 & 0.479 & 0.464 & 0.384 & 0.402 & 0.386 & 0.400 & 0.654 & 0.599 & \textbf{0.376} & 0.419 & 0.513 & 0.491 & 0.449 & 0.459 \\
 & \multicolumn{1}{c|}{192} & \underline{0.427} & \multicolumn{1}{c|}{\textbf{0.424}} & 0.443 & \multicolumn{1}{c|}{0.432} & 0.441 & 0.436 & 0.437 & \textbf{0.424} & 0.460 & 0.445 & 0.471 & 0.474 & 0.525 & 0.492 & 0.436 & 0.429 & 0.437 & 0.432 & 0.719 & 0.631 & \textbf{0.420} & 0.448 & 0.534 & 0.504 & 0.500 & 0.482 \\
 & \multicolumn{1}{c|}{336} & \underline{0.461} & \multicolumn{1}{c|}{\textbf{0.442}} & 0.463 & \multicolumn{1}{c|}{0.451} & 0.487 & 0.458 & 0.479 & \underline{0.446} & 0.501 & 0.466 & 0.570 & 0.546 & 0.565 & 0.515 & 0.491 & 0.469 & 0.481 & 0.459 & 0.778 & 0.659 & \textbf{0.459} & 0.465 & 0.588 & 0.535 & 0.521 & 0.496 \\
 & \multicolumn{1}{c|}{720} & \textbf{0.474} & \multicolumn{1}{c|}{\textbf{0.459}} & \underline{0.477} & \multicolumn{1}{c|}{\underline{0.466}} & 0.503 & 0.491 & 0.481 & 0.470 & 0.500 & 0.488 & 0.653 & 0.621 & 0.594 & 0.558 & 0.521 & 0.500 & 0.519 & 0.516 & 0.836 & 0.699 & 0.506 & 0.507 & 0.643 & 0.616 & 0.514 & 0.512 \\
 & \multicolumn{1}{c|}{AVG} & \textbf{0.435} & \multicolumn{1}{c|}{\textbf{0.430}} & \underline{0.440} & \multicolumn{1}{c|}{0.438} & 0.454 & 0.448 & 0.446 & \underline{0.434} & 0.469 & 0.454 & 0.529 & 0.522 & 0.541 & 0.507 & 0.458 & 0.450 & 0.456 & 0.452 & 0.747 & 0.647 & \underline{0.440} & 0.460 & 0.570 & 0.536 & 0.496 & 0.487 \\
 & \multicolumn{1}{c|}{} &  & \multicolumn{1}{c|}{} &  & \multicolumn{1}{c|}{} &  &  &  &  &  &  &  &  &  &  &  &  &  &  &  &  &  &  &  &  &  &  \\
\multirow{5}{*}{\rotatebox{90}{\small{ETTh2}}} & \multicolumn{1}{c|}{96} & \textbf{0.285} & \multicolumn{1}{c|}{\textbf{0.331}} & 0.296 & \multicolumn{1}{c|}{0.345} & 0.297 & 0.349 & \underline{0.288} & \underline{0.338} & 0.302 & 0.348 & 0.745 & 0.584 & 0.400 & 0.440 & 0.340 & 0.374 & 0.333 & 0.387 & 0.707 & 0.621 & 0.358 & 0.397 & 0.476 & 0.458 & 0.346 & 0.388 \\
 & \multicolumn{1}{c|}{192} & \textbf{0.368} & \multicolumn{1}{c|}{\textbf{0.384}} & 0.375 & \multicolumn{1}{c|}{0.395} & 0.380 & 0.400 & \underline{0.374} & \underline{0.390} & 0.388 & 0.400 & 0.877 & 0.656 & 0.528 & 0.509 & 0.402 & 0.414 & 0.477 & 0.476 & 0.860 & 0.689 & 0.429 & 0.439 & 0.512 & 0.493 & 0.456 & 0.452 \\
 & \multicolumn{1}{c|}{336} & \underline{0.415} & \multicolumn{1}{c|}{\underline{0.428}} & \textbf{0.414} & \multicolumn{1}{c|}{0.429} & 0.428 & 0.432 & \underline{0.415} & \textbf{0.426} & 0.426 & 0.433 & 1.043 & 0.731 & 0.643 & 0.571 & 0.452 & 0.452 & 0.594 & 0.541 & 1.000 & 0.744 & 0.496 & 0.487 & 0.552 & 0.551 & 0.482 & 0.486 \\
 & \multicolumn{1}{c|}{720} & \underline{0.419} & \multicolumn{1}{c|}{\textbf{0.438}} & \textbf{0.417} & \multicolumn{1}{c|}{0.442} & 0.427 & 0.445 & 0.420 & \underline{0.440} & 0.431 & 0.446 & 1.104 & 0.763 & 0.874 & 0.679 & 0.462 & 0.468 & 0.831 & 0.657 & 1.249 & 0.838 & 0.463 & 0.474 & 0.562 & 0.560 & 0.515 & 0.511 \\
 & \multicolumn{1}{c|}{AVG} & \textbf{0.372} & \multicolumn{1}{c|}{\textbf{0.395}} & 0.376 & \multicolumn{1}{c|}{0.403} & 0.383 & 0.406 & \underline{0.374} & \underline{0.398} & 0.387 & 0.407 & 0.942 & 0.684 & 0.611 & 0.550 & 0.414 & 0.427 & 0.559 & 0.515 & 0.954 & 0.723 & 0.436 & 0.449 & 0.526 & 0.516 & 0.450 & 0.459 \\
 & \multicolumn{1}{c|}{} &  & \multicolumn{1}{c|}{} &  & \multicolumn{1}{c|}{} &  &  &  &  &  &  &  &  &  &  &  &  &  &  &  &  &  &  &  &  &  &  \\
\multirow{5}{*}{\rotatebox{90}{\small{Weather}}} & \multicolumn{1}{c|}{96} & 0.161 & \multicolumn{1}{c|}{\underline{0.202}} & \textbf{0.153} & \multicolumn{1}{c|}{\textbf{0.200}} & 0.174 & 0.214 & 0.192 & 0.232 & 0.177 & 0.218 & \underline{0.158} & 0.230 & 0.202 & 0.261 & 0.172 & 0.220 & 0.196 & 0.255 & 0.221 & 0.306 & 0.217 & 0.296 & 0.173 & 0.223 & 0.266 & 0.336 \\
 & \multicolumn{1}{c|}{192} & 0.207 & \multicolumn{1}{c|}{\textbf{0.245}} & \textbf{0.202} & \multicolumn{1}{c|}{\textbf{0.245}} & 0.221 & 0.254 & 0.240 & 0.271 & 0.225 & 0.259 & \underline{0.206} & 0.277 & 0.242 & 0.298 & 0.219 & 0.261 & 0.237 & 0.296 & 0.261 & 0.340 & 0.276 & 0.336 & 0.245 & 0.285 & 0.307 & 0.367 \\
 & \multicolumn{1}{c|}{336} & \textbf{0.261} & \multicolumn{1}{c|}{\underline{0.285}} & \underline{0.264} & \multicolumn{1}{c|}{\textbf{0.268}} & 0.278 & 0.296 & 0.292 & 0.307 & 0.278 & 0.297 & 0.272 & 0.335 & 0.287 & 0.335 & 0.280 & 0.306 & 0.283 & 0.335 & 0.309 & 0.378 & 0.339 & 0.380 & 0.321 & 0.338 & 0.359 & 0.395 \\
 & \multicolumn{1}{c|}{720} & \underline{0.343} & \multicolumn{1}{c|}{\textbf{0.339}} & \textbf{0.341} & \multicolumn{1}{c|}{\textbf{0.339}} & 0.358 & 0.347 & 0.364 & 0.353 & 0.354 & 0.348 & 0.398 & 0.418 & 0.351 & 0.386 & 0.365 & 0.359 & 0.345 & 0.381 & 0.377 & 0.427 & 0.403 & 0.428 & 0.414 & 0.410 & 0.419 & 0.428 \\
 & \multicolumn{1}{c|}{AVG} & \underline{0.243} & \multicolumn{1}{c|}{\underline{0.268}} & \textbf{0.240} & \multicolumn{1}{c|}{\textbf{0.263}} & 0.258 & 0.278 & 0.272 & 0.291 & 0.258 & 0.280 & 0.258 & 0.315 & 0.270 & 0.320 & 0.259 & 0.286 & 0.265 & 0.317 & 0.292 & 0.363 & 0.309 & 0.360 & 0.288 & 0.314 & 0.338 & 0.382 \\
 & \multicolumn{1}{c|}{} &  & \multicolumn{1}{c|}{} &  & \multicolumn{1}{c|}{} &  &  &  &  &  &  &  &  &  &  &  &  &  &  &  &  &  &  &  &  &  &  \\
\multirow{5}{*}{\rotatebox{90}{\small{Exchange}}} & \multicolumn{1}{c|}{96} & \textbf{0.080} & \multicolumn{1}{c|}{\textbf{0.198}} & \underline{0.084} & \multicolumn{1}{c|}{\underline{0.203}} & 0.086 & 0.206 & 0.093 & 0.217 & 0.088 & 0.205 & 0.256 & 0.367 & 0.094 & 0.218 & 0.107 & 0.234 & 0.088 & 0.218 & 0.267 & 0.396 & 0.148 & 0.278 & 0.111 & 0.237 & 0.197 & 0.323 \\
 & \multicolumn{1}{c|}{192} & \textbf{0.166} & \multicolumn{1}{c|}{\textbf{0.291}} & \underline{0.172} & \multicolumn{1}{c|}{\underline{0.294}} & 0.177 & 0.299 & 0.184 & 0.307 & 0.176 & 0.299 & 0.470 & 0.509 & 0.184 & 0.307 & 0.226 & 0.344 & 0.176 & 0.315 & 0.351 & 0.459 & 0.271 & 0.315 & 0.219 & 0.335 & 0.300 & 0.369 \\
 & \multicolumn{1}{c|}{336} & \underline{0.313} & \multicolumn{1}{c|}{\underline{0.403}} & 0.314 & \multicolumn{1}{c|}{0.405} & 0.331 & 0.417 & 0.351 & 0.432 & \textbf{0.301} & \textbf{0.397} & 1.268 & 0.883 & 0.349 & 0.431 & 0.367 & 0.448 & \underline{0.313} & 0.427 & 1.324 & 0.853 & 0.460 & 0.427 & 0.421 & 0.476 & 0.509 & 0.524 \\
 & \multicolumn{1}{c|}{720} & 0.843 & \multicolumn{1}{c|}{0.692} & \textbf{0.781} & \multicolumn{1}{c|}{\textbf{0.656}} & 0.847 & \underline{0.691} & 0.886 & 0.714 & 0.901 & 0.714 & 1.767 & 1.068 & 0.852 & 0.698 & 0.964 & 0.746 & \underline{0.839} & 0.695 & 1.058 & 0.797 & 1.195 & 0.695 & 1.092 & 0.769 & 1.447 & 0.941 \\
 & \multicolumn{1}{c|}{AVG} & \underline{0.350} & \multicolumn{1}{c|}{\underline{0.396}} & \textbf{0.338} & \multicolumn{1}{c|}{\textbf{0.390}} & 0.360 & 0.403 & 0.378 & 0.418 & 0.366 & 0.404 & 0.940 & 0.707 & 0.370 & 0.414 & 0.416 & 0.443 & 0.354 & 0.414 & 0.750 & 0.626 & 0.518 & 0.429 & 0.461 & 0.454 & 0.613 & 0.539 \\ \cmidrule(l){2-28} 
\end{tabular}
}
\end{table}

\paragraph{Multiple random runs}
To evaluate the significance of our forecasting experiments, we provide the mean and standard deviation (STD) for two exemplary datasets, ETTh2 and Weather, in Table~\ref{appendix:tab:forecasting_signi}. To calculate the mean and the STD  for MSE and MAE, we performed all runs five times with different seeds. 

\begin{table}[]
\caption{Mean and STD of MSE and MAE across 5 runs for the ETTh2 and Weather datasets in the format $\texttt{Mean}\pm\texttt{STD}$.}
\label{appendix:tab:forecasting_signi}
\centering
\scalebox{.85}{
\begin{tabular}{@{}lll@{}}
\toprule
Dataset & MSE & MAE \\ \midrule
ETTh2 & 0.289$\pm$0.028 & 0.336$\pm$0.018 \\
Weather & 0.162$\pm$0.013 & 0.204$\pm$0.016 \\ \bottomrule
\end{tabular}
}
\end{table}

\section{Experimental Details}\label{app:exp_details}
This section gives additional details about our forecasting and imputation-related experiments. All experiments were part of a hyperparameter study utilizing a random search methodology. We examined the subsequent hyperparameters within these specified ranges: 
\begin{itemize}
    \item Number of \EncodingLayer{}s: $N \in [0,1,\dots12]$
    \item Number of heads in the self-attention module: $h\in\{2,4,8,16,32\}$
    \item Feature embedding size: $d\in \{8, 16, 32, 64, 128\}$
    \item Attention function: $attention\_func\in \{$classic (vanilla)~\cite{vaswani2017attention}, sparse-attention (entmax15)~\cite{wu2020adversarial}$\}$.
    \item Amount and configuration of the CNN layers in the \RepresentationLayer{}: We varied the amount of CNN layers between 1 and 4. The configuration was designed such that the smallest kernel covered around 3 values and the biggest around 50\% - 80\% of the input sequence. The kernels in between covered middle sized sequences. We also experimented with different amounts of groups in the CNN layers, as well as with depthwise-convolution to further decrease the memory footprint. 

\end{itemize}

\section{Imputation}\label{app:imputation:results}

We extend our evaluation from Section~\ref{sec:exp:imputation} with a more detailed evaluation, including more SOTA approaches for the comparison, and we provide the results of all missing ratios ($r_m$) separately (see Table~\ref{app:tab:imputation1}).
For the evaluation, we consider six datasets(i.e., ECL, ETTm1, ETTm2, ETTh1, ETTh2, Weather) and compare against multiple SOTA techniques: TimesNet~\citep{wu2022timesnet}, ETSformer~\citep{woo2022etsformer}, LightTS~\citep{zhang2022less}, DLinear~\citep{zeng2022transformers}, FEDformer~\cite{zhou2022fedformer}, Stationary~\citep{liu2022non}, Autoformer~\citep{wu2021autoformer}, Pyraformer~\citep{liu2021pyraformer}, Informer~\citep{zhou2021informer}, and LogTrans~\cite{li2019enhancing}.

\begin{table}[]
\caption{Performance comparison for the multivariate imputation task with missing ratios $r_m \in \{0.125, 0.25, 0.375, 0.5\}$ and a fixed input length of 96. Bold/underline indicate best/second.}
\label{app:tab:imputation1}
\centering
\scalebox{.53}{
\begin{tabular}{@{}cccccccccccccccccccccccccc@{}}
\toprule
\multicolumn{2}{c}{Models} & \multicolumn{2}{c}{TSRM} & \multicolumn{2}{c}{TSRM\_IFC} & \multicolumn{2}{c}{TimesNet} & \multicolumn{2}{c}{ETSformer} & \multicolumn{2}{c}{LigthTS} & \multicolumn{2}{c}{Dlinear} & \multicolumn{2}{c}{FEDformer} & \multicolumn{2}{c}{Stationary} & \multicolumn{2}{c}{Autoformer} & \multicolumn{2}{c}{Pyraformer} & \multicolumn{2}{c}{Informer} & \multicolumn{2}{c}{LogTrans} \\ \midrule
\multicolumn{2}{c|}{Metrics} & MSE & \multicolumn{1}{c|}{MAE} & MSE & \multicolumn{1}{c|}{MAE} & MSE & MAE & MSE & MAE & MSE & MAE & MSE & MAE & MSE & MAE & MSE & MAE & MSE & MAE & MSE & MAE & MSE & MAE & MSE & MAE \\
\multirow{5}{*}{\rotatebox{90}{\small{ECL}}} & \multicolumn{1}{c|}{0.125} & \textbf{0.057} & \multicolumn{1}{c|}{\textbf{0.146}} & \underline{0.060} & \multicolumn{1}{c|}{\underline{0.166}} & 0.085 & 0.202 & 0.196 & 0.321 & 0.102 & 0.229 & 0.092 & 0.214 & 0.107 & 0.237 & 0.093 & 0.210 & 0.089 & 0.210 & 0.297 & 0.383 & 0.218 & 0.326 & 0.164 & 0.296 \\
 & \multicolumn{1}{c|}{0.250} & \textbf{0.065} & \multicolumn{1}{c|}{\textbf{0.166}} & \underline{0.067} & \multicolumn{1}{c|}{\underline{0.168}} & 0.089 & 0.206 & 0.207 & 0.332 & 0.121 & 0.252 & 0.118 & 0.247 & 0.120 & 0.251 & 0.097 & 0.214 & 0.096 & 0.220 & 0.294 & 0.380 & 0.219 & 0.326 & 0.169 & 0.299 \\
 & \multicolumn{1}{c|}{0.375} & \textbf{0.072} & \multicolumn{1}{c|}{\textbf{0.166}} & \underline{0.083} & \multicolumn{1}{c|}{\underline{0.191}} & 0.094 & 0.213 & 0.219 & 0.344 & 0.141 & 0.273 & 0.144 & 0.276 & 0.136 & 0.266 & 0.102 & 0.220 & 0.104 & 0.229 & 0.296 & 0.381 & 0.222 & 0.328 & 0.178 & 0.305 \\
 & \multicolumn{1}{c|}{0.500} & \underline{0.098} & \multicolumn{1}{c|}{\underline{0.203}} & \textbf{0.082} & \multicolumn{1}{c|}{\textbf{0.191}} & 0.100 & 0.221 & 0.235 & 0.357 & 0.160 & 0.293 & 0.175 & 0.305 & 0.158 & 0.284 & 0.108 & 0.228 & 0.113 & 0.239 & 0.299 & 0.383 & 0.228 & 0.331 & 0.187 & 0.312 \\
 & \multicolumn{1}{c|}{AVG} & \textbf{0.073} & \multicolumn{1}{c|}{\textbf{0.170}} & \textbf{0.073} & \multicolumn{1}{c|}{\underline{0.179}} & 0.092 & 0.210 & 0.214 & 0.338 & 0.131 & 0.262 & 0.132 & 0.260 & 0.130 & 0.260 & 0.100 & 0.218 & 0.100 & 0.224 & 0.296 & 0.382 & 0.222 & 0.328 & 0.174 & 0.303 \\
 & \multicolumn{1}{c|}{} &  & \multicolumn{1}{c|}{} &  & \multicolumn{1}{c|}{} &  &  &  &  &  &  &  &  &  &  &  &  &  &  &  &  &  &  &  &  \\
\multirow{5}{*}{\rotatebox{90}{\small{ETTm1}}} & \multicolumn{1}{c|}{0.125} & 0.037 & \multicolumn{1}{c|}{\textbf{0.068}} & 0.031 & \multicolumn{1}{c|}{0.119} & \textbf{0.019} & \underline{0.092} & 0.067 & 0.188 & 0.075 & 0.180 & 0.058 & 0.162 & 0.035 & 0.135 & \underline{0.026} & 0.107 & 0.034 & 0.124 & 0.670 & 0.541 & 0.047 & 0.155 & 0.041 & 0.141 \\
 & \multicolumn{1}{c|}{0.250} & 0.039 & \multicolumn{1}{c|}{0.123} & 0.034 & \multicolumn{1}{c|}{0.121} & \textbf{0.023} & \textbf{0.101} & 0.096 & 0.229 & 0.093 & 0.206 & 0.080 & 0.193 & 0.052 & 0.166 & \underline{0.032} & \underline{0.119} & 0.046 & 0.144 & 0.689 & 0.553 & 0.063 & 0.180 & 0.044 & 0.144 \\
 & \multicolumn{1}{c|}{0.375} & 0.043 & \multicolumn{1}{c|}{\underline{0.130}} & 0.042 & \multicolumn{1}{c|}{0.134} & \textbf{0.029} & \textbf{0.111} & 0.133 & 0.271 & 0.113 & 0.231 & 0.103 & 0.219 & 0.069 & 0.191 & \underline{0.039} & 0.131 & 0.057 & 0.161 & 0.737 & 0.581 & 0.079 & 0.200 & 0.052 & 0.158 \\
 & \multicolumn{1}{c|}{0.500} & 0.051 & \multicolumn{1}{c|}{\underline{0.142}} & 0.072 & \multicolumn{1}{c|}{0.185} & \textbf{0.036} & \textbf{0.124} & 0.186 & 0.323 & 0.134 & 0.255 & 0.132 & 0.248 & 0.089 & 0.218 & \underline{0.047} & 0.145 & 0.067 & 0.174 & 0.770 & 0.605 & 0.093 & 0.218 & 0.063 & 0.173 \\
 & \multicolumn{1}{c|}{AVG} & 0.042 & \multicolumn{1}{c|}{\underline{0.116}} & 0.045 & \multicolumn{1}{c|}{0.140} & \textbf{0.027} & \textbf{0.107} & 0.120 & 0.253 & 0.104 & 0.218 & 0.093 & 0.206 & 0.061 & 0.178 & \underline{0.036} & 0.126 & 0.051 & 0.151 & 0.716 & 0.570 & 0.070 & 0.188 & 0.050 & 0.154 \\
 & \multicolumn{1}{c|}{} &  & \multicolumn{1}{c|}{} &  & \multicolumn{1}{c|}{} &  &  &  &  &  &  &  &  &  &  &  &  &  &  &  &  &  &  &  &  \\
\multirow{5}{*}{\rotatebox{90}{\small{ETTm2}}} & \multicolumn{1}{c|}{0.125} & 0.025 & \multicolumn{1}{c|}{0.091} & \underline{0.020} & \multicolumn{1}{c|}{\underline{0.087}} & \textbf{0.018} & \textbf{0.080} & 0.108 & 0.239 & 0.034 & 0.127 & 0.062 & 0.166 & 0.056 & 0.159 & 0.021 & 0.088 & 0.023 & 0.092 & 0.394 & 0.470 & 0.133 & 0.270 & 0.103 & 0.229 \\
 & \multicolumn{1}{c|}{0.250} & 0.027 & \multicolumn{1}{c|}{\underline{0.095}} & \underline{0.021} & \multicolumn{1}{c|}{0.096} & \textbf{0.020} & \textbf{0.085} & 0.164 & 0.294 & 0.042 & 0.143 & 0.085 & 0.196 & 0.080 & 0.195 & 0.024 & 0.096 & 0.026 & 0.101 & 0.421 & 0.482 & 0.135 & 0.272 & 0.120 & 0.248 \\
 & \multicolumn{1}{c|}{0.375} & 0.029 & \multicolumn{1}{c|}{0.109} & \textbf{0.020} & \multicolumn{1}{c|}{\underline{0.092}} & \underline{0.023} & \textbf{0.091} & 0.237 & 0.356 & 0.051 & 0.159 & 0.106 & 0.222 & 0.110 & 0.231 & 0.027 & 0.103 & 0.030 & 0.108 & 0.478 & 0.521 & 0.155 & 0.293 & 0.138 & 0.260 \\
 & \multicolumn{1}{c|}{0.500} & 0.033 & \multicolumn{1}{c|}{0.118} & 0.034 & \multicolumn{1}{c|}{0.124} & \textbf{0.026} & \textbf{0.098} & 0.323 & 0.421 & 0.059 & 0.174 & 0.131 & 0.247 & 0.156 & 0.276 & \underline{0.030} & \underline{0.108} & 0.035 & 0.119 & 0.568 & 0.560 & 0.200 & 0.333 & 0.117 & 0.247 \\
 & \multicolumn{1}{c|}{AVG} & 0.028 & \multicolumn{1}{c|}{0.103} & \underline{0.024} & \multicolumn{1}{c|}{0.100} & \textbf{0.022} & \textbf{0.088} & 0.208 & 0.328 & 0.046 & 0.151 & 0.096 & 0.208 & 0.100 & 0.215 & 0.026 & \underline{0.099} & 0.028 & 0.105 & 0.465 & 0.508 & 0.156 & 0.292 & 0.120 & 0.246 \\
 & \multicolumn{1}{c|}{} &  & \multicolumn{1}{c|}{} &  & \multicolumn{1}{c|}{} &  &  &  &  &  &  &  &  &  &  &  &  &  &  &  &  &  &  &  &  \\
\multirow{5}{*}{\rotatebox{90}{\small{ETTh1}}} & \multicolumn{1}{c|}{0.125} & 0.088 & \multicolumn{1}{c|}{0.197} & \textbf{0.046} & \multicolumn{1}{c|}{\textbf{0.146}} & \underline{0.057} & \underline{0.159} & 0.126 & 0.263 & 0.240 & 0.345 & 0.151 & 0.267 & 0.070 & 0.190 & 0.060 & 0.165 & 0.074 & 0.182 & 0.857 & 0.609 & 0.114 & 0.234 & 0.229 & 0.330 \\
 & \multicolumn{1}{c|}{0.250} & 0.092 & \multicolumn{1}{c|}{0.201} & \textbf{0.052} & \multicolumn{1}{c|}{\textbf{0.158}} & \underline{0.069} & \underline{0.178} & 0.169 & 0.304 & 0.265 & 0.364 & 0.180 & 0.292 & 0.106 & 0.236 & 0.080 & 0.189 & 0.090 & 0.203 & 0.829 & 0.672 & 0.140 & 0.262 & 0.207 & 0.323 \\
 & \multicolumn{1}{c|}{0.375} & 0.112 & \multicolumn{1}{c|}{0.221} & \textbf{0.066} & \multicolumn{1}{c|}{\textbf{0.173}} & \underline{0.084} & \underline{0.196} & 0.220 & 0.347 & 0.296 & 0.382 & 0.215 & 0.318 & 0.124 & 0.258 & 0.102 & 0.212 & 0.109 & 0.222 & 0.830 & 0.675 & 0.174 & 0.293 & 0.210 & 0.328 \\
 & \multicolumn{1}{c|}{0.500} & 0.115 & \multicolumn{1}{c|}{0.230} & \textbf{0.073} & \multicolumn{1}{c|}{\textbf{0.183}} & \underline{0.102} & \underline{0.215} & 0.293 & 0.402 & 0.334 & 0.404 & 0.257 & 0.347 & 0.165 & 0.299 & 0.133 & 0.240 & 0.137 & 0.248 & 0.854 & 0.691 & 0.215 & 0.325 & 0.230 & 0.348 \\
 & \multicolumn{1}{c|}{AVG} & 0.102 & \multicolumn{1}{c|}{0.212} & \textbf{0.059} & \multicolumn{1}{c|}{\textbf{0.165}} & \underline{0.078} & \underline{0.187} & 0.202 & 0.329 & 0.284 & 0.374 & 0.201 & 0.306 & 0.116 & 0.246 & 0.094 & 0.202 & 0.102 & 0.214 & 0.842 & 0.662 & 0.161 & 0.278 & 0.219 & 0.332 \\
 & \multicolumn{1}{c|}{} &  & \multicolumn{1}{c|}{} &  & \multicolumn{1}{c|}{} &  &  &  &  &  &  &  &  &  &  &  &  &  &  &  &  &  &  &  &  \\
\multirow{5}{*}{\rotatebox{90}{\small{ETTh2}}} & \multicolumn{1}{c|}{0.125} & 0.060 & \multicolumn{1}{c|}{0.156} & 0.045 & \multicolumn{1}{c|}{0.141} & \textbf{0.040} & \textbf{0.130} & 0.187 & 0.319 & 0.101 & 0.231 & 0.100 & 0.216 & 0.095 & 0.212 & \underline{0.042} & \underline{0.133} & 0.044 & 0.138 & 0.976 & 0.754 & 0.305 & 0.431 & 0.173 & 0.308 \\
 & \multicolumn{1}{c|}{0.250} & 0.063 & \multicolumn{1}{c|}{0.162} & \textbf{0.046} & \multicolumn{1}{c|}{0.149} & \textbf{0.046} & \textbf{0.141} & 0.279 & 0.390 & 0.115 & 0.246 & 0.127 & 0.247 & 0.137 & 0.258 & 0.049 & \underline{0.147} & 0.050 & 0.149 & 1.037 & 0.774 & 0.322 & 0.444 & 0.175 & 0.310 \\
 & \multicolumn{1}{c|}{0.375} & 0.069 & \multicolumn{1}{c|}{0.168} & \textbf{0.045} & \multicolumn{1}{c|}{\textbf{0.143}} & \underline{0.052} & \underline{0.151} & 0.400 & 0.465 & 0.126 & 0.257 & 0.158 & 0.276 & 0.187 & 0.304 & 0.056 & 0.158 & 0.060 & 0.163 & 1.107 & 0.800 & 0.353 & 0.462 & 0.185 & 0.315 \\
 & \multicolumn{1}{c|}{0.500} & 0.097 & \multicolumn{1}{c|}{0.207} & 0.068 & \multicolumn{1}{c|}{\underline{0.170}} & \textbf{0.060} & \textbf{0.162} & 0.602 & 0.572 & 0.136 & 0.268 & 0.183 & 0.299 & 0.232 & 0.341 & \underline{0.065} & \underline{0.170} & 0.068 & 0.173 & 1.193 & 0.838 & 0.369 & 0.472 & 0.212 & 0.339 \\
 & \multicolumn{1}{c|}{AVG} & 0.072 & \multicolumn{1}{c|}{0.173} & \underline{0.051} & \multicolumn{1}{c|}{\underline{0.151}} & \textbf{0.050} & \textbf{0.146} & 0.367 & 0.436 & 0.120 & 0.250 & 0.142 & 0.260 & 0.163 & 0.279 & 0.053 & 0.152 & 0.056 & 0.156 & 1.078 & 0.792 & 0.337 & 0.452 & 0.186 & 0.318 \\
 & \multicolumn{1}{c|}{} &  & \multicolumn{1}{c|}{} &  & \multicolumn{1}{c|}{} &  &  &  &  &  &  &  &  &  &  &  &  &  &  &  &  &  &  &  &  \\
\multirow{5}{*}{\rotatebox{90}{\small{Weather}}} & \multicolumn{1}{c|}{0.125} & \underline{0.025} & \multicolumn{1}{c|}{\underline{0.044}} & \textbf{0.024} & \multicolumn{1}{c|}{\textbf{0.043}} & \underline{0.025} & 0.045 & 0.057 & 0.141 & 0.047 & 0.101 & 0.039 & 0.084 & 0.041 & 0.107 & 0.027 & 0.051 & 0.026 & 0.047 & 0.140 & 0.220 & 0.037 & 0.093 & 0.037 & 0.072 \\
 & \multicolumn{1}{c|}{0.250} & \underline{0.029} & \multicolumn{1}{c|}{\underline{0.046}} & \textbf{0.026} & \multicolumn{1}{c|}{\textbf{0.040}} & \underline{0.029} & 0.052 & 0.065 & 0.155 & 0.052 & 0.111 & 0.048 & 0.103 & 0.064 & 0.163 & \underline{0.029} & 0.056 & 0.030 & 0.054 & 0.147 & 0.229 & 0.042 & 0.100 & 0.038 & 0.074 \\
 & \multicolumn{1}{c|}{0.375} & \underline{0.031} & \multicolumn{1}{c|}{\underline{0.049}} & \textbf{0.029} & \multicolumn{1}{c|}{\textbf{0.047}} & \underline{0.031} & 0.057 & 0.081 & 0.180 & 0.058 & 0.121 & 0.057 & 0.117 & 0.107 & 0.229 & 0.033 & 0.062 & 0.032 & 0.060 & 0.156 & 0.240 & 0.049 & 0.111 & 0.039 & 0.078 \\
 & \multicolumn{1}{c|}{0.500} & 0.038 & \multicolumn{1}{c|}{\underline{0.054}} & \underline{0.037} & \multicolumn{1}{c|}{\textbf{0.051}} & \textbf{0.034} & 0.062 & 0.102 & 0.207 & 0.065 & 0.133 & 0.066 & 0.134 & 0.183 & 0.312 & \underline{0.037} & 0.068 & \underline{0.037} & 0.067 & 0.164 & 0.249 & 0.053 & 0.114 & 0.042 & 0.082 \\
 & AVG & 0.031 & \underline{0.048} & \textbf{0.029} & \textbf{0.045} & \underline{0.030} & 0.054 & 0.076 & 0.171 & 0.056 & 0.116 & 0.052 & 0.110 & 0.099 & 0.203 & 0.032 & 0.059 & 0.031 & 0.057 & 0.152 & 0.234 & 0.045 & 0.104 & 0.039 & 0.076 \\ \cmidrule(l){2-26} 
\end{tabular}
}
\end{table}

\section{Ablation Study}\label{app:ablation}

In addition to the ablation study results in Section~\ref{sec:ablation}, we provide more detailed results for each experiment. Table~\ref{tab:ablation:N} shows all details regarding the sensitivity study for the parameter $N$, including MSE and MAE metrics for both datasets.
We further provide the exact results of the architecture variation study in Table~\ref{tab:ablation:variations}. Again, we provide the MSE and MAE metrics for both datasets, Weather and ETTh1.

\begin{table}[]
\caption{Hyperparameter sensitivity study for $N$.}
\label{tab:ablation:N}
\centering
\scalebox{0.7}{
\begin{tabular}{@{}ccccccccccccccccccc@{}}
\toprule
Dataset & \multicolumn{2}{c}{N=0} & \multicolumn{2}{c}{N=1} & \multicolumn{2}{c}{N=2} & \multicolumn{2}{c}{N=3} & \multicolumn{2}{c}{N=4} & \multicolumn{2}{c}{N=5} & \multicolumn{2}{c}{N=6} & \multicolumn{2}{c}{N=7} & \multicolumn{2}{c}{N=8} \\ \midrule
 & MSE & MAE & MSE & MAE & MSE & MAE & MSE & MAE & MSE & MAE & MSE & MAE & MSE & MAE & MSE & MAE & MSE & MAE \\
ETTh1 & 0.406 & 0.411 & 0.381 & 0.395 & 0.393 & 0.401 & 0.383 & \textbf{0.394} & 0.398 & 0.403 & 0.398 & 0.403 & 0.384 & 0.397 & \textbf{0.379} & \textbf{0.394} & 0.396 & 0.407 \\
Weather & 0.217 & 0.248 & 0.163 & 0.205 & 0.162 & \textbf{0.204} & 0.162 & \textbf{0.204} & \textbf{0.161} & \textbf{0.204} & 0.162 & 0.204 & 0.162 & 0.205 & 0.167 & 0.298 & 0.168 & 0.210 \\ \bottomrule
\end{tabular}
}
\end{table}

\begin{table}[]
\centering
\caption{Ablation study with different architectural variations.}
\label{tab:ablation:variations}
\scalebox{0.73}{
\begin{tabular}{@{}ccccccccccc@{}}
\toprule
Dataset & \multicolumn{2}{c}{TSRM} & \multicolumn{2}{c}{TSRM\_{IFC}} & \multicolumn{2}{c}{no\_merge} & \multicolumn{2}{c}{no\_merge+$R1$} & \multicolumn{2}{c}{no\_merge+$R0$} \\ \midrule
 & MSE & MAE & MSE & MAE & MSE & MAE & MSE & MAE & MSE & MAE \\
ETTh1 & \textbf{0.379} & \textbf{0.394} & 0.387 & 0.401 & 0.382 & \textbf{0.394} & 0.398 & 0.403 & 0.405 & 0.416 \\
Weather & 0.161 & 0.202 & \textbf{0.153} & \textbf{0.200} & 0.165 & 0.206 & 0.165 & 0.206 & 0.166 & 0.206 \\ \bottomrule
\end{tabular}
}
\end{table}

\begin{table}[]
\label{tab:fc_var_T}
\caption{Performance comparison for the multivariate forecasting task with forecasting horizons $H \in \{96, 192, 336, 720\}$ and variable look-back window $T \in \{96, 192, 336, 720\}$. AVG shows the averaged result over all forecast horizons per data set and model. Bold/underlined is the best/second result.}
\centering

\begin{tabular}{@{}cccccccccccccc@{}}

\toprule
\multicolumn{2}{c}{Models} & \multicolumn{2}{c}{TSRM} & \multicolumn{2}{c}{TSRM\_M} & \multicolumn{2}{c}{PatchTST} & \multicolumn{2}{c}{DLinear} & \multicolumn{2}{c}{FEDformer} & \multicolumn{2}{c}{Autoformer} \\ \midrule
\multicolumn{2}{c}{Metrics} & MSE & \multicolumn{1}{c|}{MAE} & MSE & \multicolumn{1}{c|}{MAE} & MSE & MAE & MSE & MAE & MSE & MAE & MSE & MAE \\
 & \multicolumn{1}{c|}{} &  & \multicolumn{1}{c|}{} &  & \multicolumn{1}{c|}{} &  &  &  &  &  &  &  &  \\
\multirow{5}{*}{\rotatebox{90}{\small{ECL}}} & \multicolumn{1}{c|}{96} & \underline{0.132} & \multicolumn{1}{c|}{0.255} & 0.148 & \multicolumn{1}{c|}{0.247} & \textbf{0.129} & \textbf{0.222} & 0.140 & \underline{0.237} & 0.186 & 0.302 & 0.196 & 0.313 \\
 & \multicolumn{1}{c|}{192} & \underline{0.148} & \multicolumn{1}{c|}{\textbf{0.240}} & 0.168 & \multicolumn{1}{c|}{0.263} & \textbf{0.147} & \textbf{0.240} & 0.153 & 0.249 & 0.197 & 0.311 & 0.211 & 0.324 \\
 & \multicolumn{1}{c|}{336} & \underline{0.169} & \multicolumn{1}{c|}{\underline{0.261}} & 0.175 & \multicolumn{1}{c|}{0.274} & \textbf{0.163} & \textbf{0.259} & \underline{0.169} & 0.267 & 0.213 & 0.328 & 0.214 & 0.327 \\
 & \multicolumn{1}{c|}{720} & \underline{0.197} & \multicolumn{1}{c|}{\textbf{0.283}} & \textbf{0.196} & \multicolumn{1}{c|}{0.292} & \underline{0.197} & \underline{0.290} & 0.203 & 0.301 & 0.233 & 0.344 & 0.236 & 0.342 \\
 & \multicolumn{1}{c|}{AVG} & \underline{0.162} & \multicolumn{1}{c|}{\underline{0.260}} & 0.172 & \multicolumn{1}{c|}{0.269} & \textbf{0.159} & \textbf{0.253} & 0.166 & 0.264 & 0.207 & 0.321 & 0.214 & 0.326 \\
 & \multicolumn{1}{c|}{} &  & \multicolumn{1}{c|}{} &  & \multicolumn{1}{c|}{} &  &  &  &  &  &  &  &  \\
\multirow{5}{*}{\rotatebox{90}{\small{ETTm1}}} & \multicolumn{1}{c|}{96} & \textbf{0.290} & \multicolumn{1}{c|}{\textbf{0.339}} & 0.296 & \multicolumn{1}{c|}{\underline{0.340}} & \textbf{0.290} & 0.342 & 0.299 & 0.343 & 0.326 & 0.390 & 0.510 & 0.492 \\
 & \multicolumn{1}{c|}{192} & \textbf{0.328} & \multicolumn{1}{c|}{\textbf{0.360}} & 0.353 & \multicolumn{1}{c|}{0.381} & \underline{0.332} & 0.369 & 0.335 & \underline{0.365} & 0.365 & 0.415 & 0.514 & 0.495 \\
 & \multicolumn{1}{c|}{336} & \textbf{0.360} & \multicolumn{1}{c|}{\textbf{0.385}} & 0.389 & \multicolumn{1}{c|}{0.403} & \underline{0.366} & 0.392 & 0.369 & \underline{0.386} & 0.392 & 0.425 & 0.510 & 0.492 \\
 & \multicolumn{1}{c|}{720} & \textbf{0.407} & \multicolumn{1}{c|}{0.425} & 0.421 & \multicolumn{1}{c|}{0.435} & \underline{0.420} & \underline{0.424} & 0.425 & \textbf{0.421} & 0.446 & 0.458 & 0.527 & 0.493 \\
 & \multicolumn{1}{c|}{AVG} & \textbf{0.346} & \multicolumn{1}{c|}{\textbf{0.377}} & 0.365 & \multicolumn{1}{c|}{0.390} & \underline{0.352} & 0.382 & 0.357 & \underline{0.379} & 0.382 & 0.422 & 0.515 & 0.493 \\
 & \multicolumn{1}{c|}{} &  & \multicolumn{1}{c|}{} &  & \multicolumn{1}{c|}{} &  &  &  &  &  &  &  &  \\
\multirow{5}{*}{\rotatebox{90}{\small{ETTm2}}} & \multicolumn{1}{c|}{96} & \underline{0.164} & \multicolumn{1}{c|}{\underline{0.253}} & \textbf{0.161} & \multicolumn{1}{c|}{\textbf{0.251}} & 0.165 & 0.255 & 0.167 & 0.260 & 0.180 & 0.271 & 0.205 & 0.293 \\
 & \multicolumn{1}{c|}{192} & \textbf{0.219} & \multicolumn{1}{c|}{\textbf{0.291}} & 0.224 & \multicolumn{1}{c|}{0.296} & \underline{0.220} & \underline{0.292} & 0.224 & 0.303 & 0.252 & 0.318 & 0.278 & 0.336 \\
 & \multicolumn{1}{c|}{336} & \textbf{0.273} & \multicolumn{1}{c|}{\underline{0.328}} & 0.275 & \multicolumn{1}{c|}{\textbf{0.327}} & \underline{0.274} & 0.329 & 0.281 & 0.342 & 0.324 & 0.364 & 0.343 & 0.379 \\
 & \multicolumn{1}{c|}{720} & \textbf{0.360} & \multicolumn{1}{c|}{\textbf{0.384}} & 0.382 & \multicolumn{1}{c|}{0.392} & \underline{0.362} & \underline{0.385} & 0.397 & 0.421 & 0.410 & 0.420 & 0.414 & 0.419 \\
 & \multicolumn{1}{c|}{AVG} & \textbf{0.254} & \multicolumn{1}{c|}{\textbf{0.314}} & 0.260 & \multicolumn{1}{c|}{0.316} & \underline{0.255} & \underline{0.315} & 0.267 & 0.332 & 0.292 & 0.343 & 0.310 & 0.357 \\
 & \multicolumn{1}{c|}{} &  & \multicolumn{1}{c|}{} &  & \multicolumn{1}{c|}{} &  &  &  &  &  &  &  &  \\
\multirow{5}{*}{\rotatebox{90}{\small{ETTh1}}} & \multicolumn{1}{c|}{96} & \textbf{0.361} & \multicolumn{1}{c|}{\textbf{0.391}} & 0.385 & \multicolumn{1}{c|}{0.414} & \underline{0.370} & 0.400 & 0.375 & \underline{0.399} & 0.376 & 0.415 & 0.435 & 0.446 \\
 & \multicolumn{1}{c|}{192} & \textbf{0.404} & \multicolumn{1}{c|}{\underline{0.423}} & 0.420 & \multicolumn{1}{c|}{0.434} & 0.413 & 0.429 & \underline{0.405} & \textbf{0.416} & 0.423 & 0.446 & 0.456 & 0.457 \\
 & \multicolumn{1}{c|}{336} & \underline{0.439} & \multicolumn{1}{c|}{\textbf{0.438}} & 0.462 & \multicolumn{1}{c|}{0.449} & \textbf{0.422} & \underline{0.440} & \underline{0.439} & 0.443 & 0.444 & 0.462 & 0.486 & 0.487 \\
 & \multicolumn{1}{c|}{720} & \textbf{0.439} & \multicolumn{1}{c|}{\textbf{0.457}} & 0.477 & \multicolumn{1}{c|}{\underline{0.466}} & \underline{0.447} & 0.468 & 0.472 & 0.490 & 0.469 & 0.492 & 0.515 & 0.517 \\
 & \multicolumn{1}{c|}{AVG} & \textbf{0.411} & \multicolumn{1}{c|}{\textbf{0.427}} & 0.436 & \multicolumn{1}{c|}{0.441} & \underline{0.413} & \underline{0.434} & 0.423 & 0.437 & 0.428 & 0.454 & 0.473 & 0.477 \\
 & \multicolumn{1}{c|}{} &  & \multicolumn{1}{c|}{} &  & \multicolumn{1}{c|}{} &  &  &  &  &  &  &  &  \\
\multirow{5}{*}{\rotatebox{90}{\small{ETTh2}}} & \multicolumn{1}{c|}{96} & \underline{0.276} & \multicolumn{1}{c|}{\textbf{0.336}} & 0.286 & \multicolumn{1}{c|}{\underline{0.337}} & \textbf{0.274} & \underline{0.337} & 0.289 & 0.353 & 0.332 & 0.374 & 0.332 & 0.368 \\
 & \multicolumn{1}{c|}{192} & \underline{0.344} & \multicolumn{1}{c|}{\textbf{0.378}} & 0.350 & \multicolumn{1}{c|}{0.390} & \textbf{0.341} & \underline{0.382} & 0.383 & 0.418 & 0.407 & 0.446 & 0.426 & 0.434 \\
 & \multicolumn{1}{c|}{336} & 0.369 & \multicolumn{1}{c|}{0.412} & \underline{0.361} & \multicolumn{1}{c|}{\underline{0.409}} & \textbf{0.329} & \textbf{0.384} & 0.448 & 0.465 & 0.400 & 0.447 & 0.477 & 0.479 \\
 & \multicolumn{1}{c|}{720} & 0.402 & \multicolumn{1}{c|}{0.432} & \underline{0.399} & \multicolumn{1}{c|}{\underline{0.426}} & \textbf{0.379} & \textbf{0.422} & 0.605 & 0.551 & 0.412 & 0.469 & 0.453 & 0.490 \\
 & \multicolumn{1}{c|}{AVG} & \underline{0.348} & \multicolumn{1}{c|}{\underline{0.390}} & 0.349 & \multicolumn{1}{c|}{\underline{0.390}} & \textbf{0.331} & \textbf{0.381} & 0.431 & 0.447 & 0.388 & 0.434 & 0.422 & 0.443 \\
 & \multicolumn{1}{c|}{} &  & \multicolumn{1}{c|}{} &  & \multicolumn{1}{c|}{} &  &  &  &  &  &  &  &  \\
\multirow{5}{*}{\rotatebox{90}{\small{Weather}}} & \multicolumn{1}{c|}{96} & \textbf{0.144} & \multicolumn{1}{c|}{\textbf{0.188}} & 0.153 & \multicolumn{1}{c|}{0.200} & \underline{0.149} & \underline{0.198} & 0.176 & 0.237 & 0.238 & 0.314 & 0.249 & 0.329 \\
 & \multicolumn{1}{c|}{192} & \textbf{0.188} & \multicolumn{1}{c|}{\textbf{0.230}} & 0.202 & \multicolumn{1}{c|}{0.245} & \underline{0.194} & \underline{0.241} & 0.220 & 0.282 & 0.275 & 0.329 & 0.325 & 0.370 \\
 & \multicolumn{1}{c|}{336} & \textbf{0.239} & \multicolumn{1}{c|}{\underline{0.271}} & 0.264 & \multicolumn{1}{c|}{\textbf{0.268}} & \underline{0.245} & 0.282 & 0.265 & 0.319 & 0.339 & 0.377 & 0.351 & 0.391 \\
 & \multicolumn{1}{c|}{720} & \underline{0.315} & \multicolumn{1}{c|}{\textbf{0.325}} & 0.342 & \multicolumn{1}{c|}{0.340} & \textbf{0.314} & \underline{0.334} & 0.323 & 0.362 & 0.389 & 0.409 & 0.415 & 0.426 \\
 & \multicolumn{1}{c|}{AVG} & \textbf{0.222} & \multicolumn{1}{c|}{\textbf{0.254}} & 0.240 & \multicolumn{1}{c|}{\underline{0.263}} & \underline{0.226} & 0.264 & 0.246 & 0.300 & 0.310 & 0.357 & 0.335 & 0.379 \\ \cmidrule(l){2-14} 
\end{tabular}
\end{table}

\begin{thebibliography}{10}

\bibitem{papadimitriou2006optimal}
Spiros Papadimitriou and Philip Yu.
\newblock Optimal multi-scale patterns in time series streams.
\newblock In {\em Proceedings of the 2006 ACM SIGMOD international conference
  on Management of data}, pages 647--658, 2006.

\bibitem{zhu2002statstream}
Yunyue Zhu and Dennis Shasha.
\newblock Statstream: Statistical monitoring of thousands of data streams in
  real time.
\newblock In {\em VLDB'02: Proceedings of the 28th International Conference on
  Very Large Databases}, pages 358--369. Elsevier, 2002.

\bibitem{ek2023transformer}
Sannara Ek, Fran{\c{c}}ois Portet, and Philippe Lalanda.
\newblock Transformer-based models to deal with heterogeneous environments in
  human activity recognition.
\newblock {\em Personal and Ubiquitous Computing}, 27(6):2267--2280, 2023.

\bibitem{hochreiter2001gradient}
Sepp Hochreiter, Yoshua Bengio, Paolo Frasconi, J{\"u}rgen Schmidhuber, et~al.
\newblock Gradient flow in recurrent nets: the difficulty of learning long-term
  dependencies, 2001.

\bibitem{vaswani2017attention}
Ashish Vaswani, Noam Shazeer, Niki Parmar, Jakob Uszkoreit, Llion Jones,
  Aidan~N Gomez, {\L}ukasz Kaiser, and Illia Polosukhin.
\newblock Attention is all you need.
\newblock In {\em Advances in neural information processing systems}, pages
  5998--6008, 2017.

\bibitem{wu2020deep}
Neo Wu, Bradley Green, Xue Ben, and Shawn O'Banion.
\newblock Deep transformer models for time series forecasting: The influenza
  prevalence case.
\newblock {\em arXiv preprint arXiv:2001.08317}, 2020.

\bibitem{huang2018improved}
Cheng-Zhi~Anna Huang, Ashish Vaswani, Jakob Uszkoreit, Noam Shazeer, Curtis
  Hawthorne, Andrew~M Dai, Matthew~D Hoffman, and Douglas Eck.
\newblock An improved relative self-attention mechanism for transformer with
  application to music generation.
\newblock {\em arXiv preprint arXiv:1809.04281}, 2, 2018.

\bibitem{povey2018time}
Daniel Povey, Hossein Hadian, Pegah Ghahremani, Ke~Li, and Sanjeev Khudanpur.
\newblock A time-restricted self-attention layer for asr.
\newblock In {\em 2018 IEEE International Conference on Acoustics, Speech and
  Signal Processing (ICASSP)}, pages 5874--5878. IEEE, 2018.

\bibitem{li2019enhancing}
Shiyang Li, Xiaoyong Jin, Yao Xuan, Xiyou Zhou, Wenhu Chen, Yu-Xiang Wang, and
  Xifeng Yan.
\newblock Enhancing the locality and breaking the memory bottleneck of
  transformer on time series forecasting.
\newblock {\em Advances in neural information processing systems}, 32, 2019.

\bibitem{zhou2021informer}
Haoyi Zhou, Shanghang Zhang, Jieqi Peng, Shuai Zhang, Jianxin Li, Hui Xiong,
  and Wancai Zhang.
\newblock Informer: Beyond efficient transformer for long sequence time-series
  forecasting.
\newblock In {\em Proceedings of the AAAI conference on artificial
  intelligence}, volume~35, pages 11106--11115, 2021.

\bibitem{wu2021autoformer}
Haixu Wu, Jiehui Xu, Jianmin Wang, and Mingsheng Long.
\newblock Autoformer: Decomposition transformers with auto-correlation for
  long-term series forecasting.
\newblock {\em Advances in Neural Information Processing Systems},
  34:22419--22430, 2021.

\bibitem{zhou2022fedformer}
Tian Zhou, Ziqing Ma, Qingsong Wen, Xue Wang, Liang Sun, and Rong Jin.
\newblock Fedformer: Frequency enhanced decomposed transformer for long-term
  series forecasting.
\newblock In {\em International Conference on Machine Learning}, pages
  27268--27286. PMLR, 2022.

\bibitem{cao2018brits}
Wei Cao, Dong Wang, Jian Li, Hao Zhou, Lei Li, and Yitan Li.
\newblock Brits: Bidirectional recurrent imputation for time series.
\newblock In {\em Advances in Neural Information Processing Systems}, pages
  6775--6785, 2018.

\bibitem{fortuin2020gp}
Vincent Fortuin, Dmitry Baranchuk, Gunnar R{\"a}tsch, and Stephan Mandt.
\newblock Gp-vae: Deep probabilistic time series imputation.
\newblock In {\em International conference on artificial intelligence and
  statistics}, pages 1651--1661. PMLR, 2020.

\bibitem{luo2018multivariate}
Yonghong Luo, Xiangrui Cai, Ying Zhang, Jun Xu, et~al.
\newblock Multivariate time series imputation with generative adversarial
  networks.
\newblock {\em Advances in neural information processing systems}, 31, 2018.

\bibitem{du2023saits}
Wenjie Du, David C{\^o}t{\'e}, and Yan Liu.
\newblock Saits: Self-attention-based imputation for time series.
\newblock {\em Expert Systems with Applications}, 219:119619, 2023.

\bibitem{zeng2023transformers}
Ailing Zeng, Muxi Chen, Lei Zhang, and Qiang Xu.
\newblock Are transformers effective for time series forecasting?
\newblock In {\em Proceedings of the AAAI conference on artificial
  intelligence}, volume~37, pages 11121--11128, 2023.

\bibitem{nie2022time}
Yuqi Nie, Nam~H Nguyen, Phanwadee Sinthong, and Jayant Kalagnanam.
\newblock A time series is worth 64 words: Long-term forecasting with
  transformers.
\newblock {\em arXiv preprint arXiv:2211.14730}, 2022.

\bibitem{liu2024itransformer}
Yong Liu, Tengge Hu, Haoran Zhang, Haixu Wu, Shiyu Wang, Lintao Ma, and
  Mingsheng Long.
\newblock itransformer: Inverted transformers are effective for time series
  forecasting.
\newblock {\em arXiv preprint arXiv:2310.06625}, 2024.

\bibitem{chen2024pathformer}
Peng Chen, Yingying Zhang, Yunyao Cheng, Yang Shu, Yihang Wang, Qingsong Wen,
  Bin Yang, and Chenjuan Guo.
\newblock Pathformer: Multi-scale transformers with adaptive pathways for time
  series forecasting.
\newblock In {\em International Conference on Learning Representations}, 2024.

\bibitem{wu2022timesnet}
Haixu Wu, Tengge Hu, Yong Liu, Hang Zhou, Jianmin Wang, and Mingsheng Long.
\newblock Timesnet: Temporal 2d-variation modeling for general time series
  analysis.
\newblock {\em arXiv preprint arXiv:2210.02186}, 2022.

\bibitem{kim2021reversible}
Taesung Kim, Jinhee Kim, Yunwon Tae, Cheonbok Park, Jang-Ho Choi, and Jaegul
  Choo.
\newblock Reversible instance normalization for accurate time-series
  forecasting against distribution shift.
\newblock In {\em International Conference on Learning Representations}, 2021.

\bibitem{he2016deep}
Kaiming He, Xiangyu Zhang, Shaoqing Ren, and Jian Sun.
\newblock Deep residual learning for image recognition.
\newblock In {\em Proceedings of the IEEE conference on computer vision and
  pattern recognition}, pages 770--778, 2016.

\bibitem{das2023long}
Abhimanyu Das, Weihao Kong, Andrew Leach, Shaan Mathur, Rajat Sen, and Rose Yu.
\newblock Long-term forecasting with tide: Time-series dense encoder.
\newblock {\em arXiv preprint arXiv:2304.08424}, 2023.

\bibitem{selva2021review}
S~Selva~Birunda and R~Kanniga~Devi.
\newblock A review on word embedding techniques for text classification.
\newblock {\em Innovative Data Communication Technologies and Application:
  Proceedings of ICIDCA 2020}, pages 267--281, 2021.

\bibitem{he2016identity}
Kaiming He, Xiangyu Zhang, Shaoqing Ren, and Jian Sun.
\newblock Identity mappings in deep residual networks.
\newblock In {\em ECCV}, October 2016.

\bibitem{wu2020adversarial}
Sifan Wu, Xi~Xiao, Qianggang Ding, Peilin Zhao, Ying Wei, and Junzhou Huang.
\newblock Adversarial sparse transformer for time series forecasting.
\newblock {\em Advances in Neural Information Processing Systems}, 33, 2020.

\bibitem{dua2017uci}
Dheeru Dua, Casey Graff, et~al.
\newblock Uci machine learning repository.
\newblock 2017.

\bibitem{li2023revisiting}
Zhe Li, Shiyi Qi, Yiduo Li, and Zenglin Xu.
\newblock Revisiting long-term time series forecasting: An investigation on
  linear mapping.
\newblock {\em arXiv preprint arXiv:2305.10721}, 2023.

\bibitem{zhang2022less}
Tianping Zhang, Yizhuo Zhang, Wei Cao, Jiang Bian, Xiaohan Yi, Shun Zheng, and
  Jian Li.
\newblock Less is more: Fast multivariate time series forecasting with light
  sampling-oriented mlp structures.
\newblock {\em arXiv preprint arXiv:2207.01186}, 2022.

\bibitem{zeng2022transformers}
Ailing Zeng, Muxi Chen, Lei Zhang, and Qiang Xu.
\newblock Are transformers effective for time series forecasting?
\newblock {\em arXiv preprint arXiv:2205.13504}, 2022.

\bibitem{liu2022non}
Yong Liu, Haixu Wu, Jianmin Wang, and Mingsheng Long.
\newblock Non-stationary transformers: Exploring the stationarity in time
  series forecasting.
\newblock {\em Advances in Neural Information Processing Systems},
  35:9881--9893, 2022.

\bibitem{wang2024tssurvey}
Yuxuan Wang, Haixu Wu, Jiaxiang Dong, Yong Liu, Mingsheng Long, and Jianmin
  Wang.
\newblock Deep time series models: A comprehensive survey and benchmark.
\newblock 2024.

\bibitem{zhang2023crossformer}
Yunhao Zhang and Junchi Yan.
\newblock Crossformer: Transformer utilizing cross-dimension dependency for
  multivariate time series forecasting.
\newblock In {\em The eleventh international conference on learning
  representations}, 2023.

\bibitem{zhao2024himtm}
Shubao Zhao, Ming Jin, Zhaoxiang Hou, Chengyi Yang, Zengxiang Li, Qingsong Wen,
  and Yi~Wang.
\newblock Himtm: Hierarchical multi-scale masked time series modeling for
  long-term forecasting.
\newblock {\em arXiv preprint arXiv:2401.05012}, 2024.

\bibitem{das2024decoder}
Abhimanyu Das, Weihao Kong, Rajat Sen, and Yichen Zhou.
\newblock A decoder-only foundation model for time-series forecasting.
\newblock {\em arXiv preprint arXiv:2310.10688}, 2024.

\bibitem{jiang2022transferability}
Junguang Jiang, Yang Shu, Jianmin Wang, and Mingsheng Long.
\newblock Transferability in deep learning: A survey.
\newblock {\em arXiv preprint arXiv:2201.05867}, 2022.

\bibitem{ekambaram2023tsmixer}
Vijay Ekambaram, Arindam Jati, Nam Nguyen, Phanwadee Sinthong, and Jayant
  Kalagnanam.
\newblock Tsmixer: Lightweight mlp-mixer model for multivariate time series
  forecasting.
\newblock In {\em Proceedings of the 29th ACM SIGKDD Conference on Knowledge
  Discovery and Data Mining}, page 459–469, New York, NY, USA, 2023.
  Association for Computing Machinery.

\bibitem{dong2023simmtm}
Jiaxiang Dong, Haixu Wu, Haoran Zhang, Li~Zhang, Jianmin Wang, and Mingsheng
  Long.
\newblock Simmtm: A simple pre-training framework for masked time-series
  modeling.
\newblock In A.~Oh, T.~Naumann, A.~Globerson, K.~Saenko, M.~Hardt, and
  S.~Levine, editors, {\em Advances in Neural Information Processing Systems},
  volume~36, pages 29996--30025. Curran Associates, Inc., 2023.

\bibitem{woo2022cost}
Gerald Woo, Chenghao Liu, Doyen Sahoo, Akshat Kumar, and Steven Hoi.
\newblock Co{ST}: Contrastive learning of disentangled seasonal-trend
  representations for time series forecasting.
\newblock In {\em International Conference on Learning Representations}, 2022.

\bibitem{lee2024learning}
Seunghan Lee, Taeyoung Park, and Kibok Lee.
\newblock Learning to embed time series patches independently.
\newblock In {\em The International Conference on Learning Representations
  ({ICLR})}, 2024.

\bibitem{bommasani2021opportunities}
Rishi Bommasani, Drew~A Hudson, Ehsan Adeli, Russ Altman, Simran Arora, Sydney
  von Arx, Michael~S Bernstein, Jeannette Bohg, Antoine Bosselut, Emma
  Brunskill, et~al.
\newblock On the opportunities and risks of foundation models.
\newblock {\em arXiv preprint arXiv:2108.07258}, 2021.

\bibitem{zhang2022self}
Xiang Zhang, Ziyuan Zhao, Theodoros Tsiligkaridis, and Marinka Zitnik.
\newblock Self-supervised contrastive pre-training for time series via
  time-frequency consistency.
\newblock {\em Advances in Neural Information Processing Systems},
  35:3988--4003, 2022.

\bibitem{zhou2023one}
Tian Zhou, Peisong Niu, Liang Sun, Rong Jin, et~al.
\newblock One fits all: Power general time series analysis by pretrained lm.
\newblock {\em Advances in neural information processing systems},
  36:43322--43355, 2023.

\bibitem{rasul2023lag}
Kashif Rasul, Arjun Ashok, Andrew~Robert Williams, Arian Khorasani, George
  Adamopoulos, Rishika Bhagwatkar, Marin Bilo{\v{s}}, Hena Ghonia,
  Nadhir~Vincent Hassen, Anderson Schneider, et~al.
\newblock Lag-llama: Towards foundation models for time series forecasting.
\newblock {\em arXiv preprint arXiv:2310.08278}, 2023.

\bibitem{goswami2024moment}
Mononito Goswami, Konrad Szafer, Arjun Choudhry, Yifu Cai, Shuo Li, and Artur
  Dubrawski.
\newblock {MOMENT}: A family of open time-series foundation models.
\newblock In {\em Forty-first International Conference on Machine Learning},
  2024.

\bibitem{woo2024unified}
Gerald Woo, Chenghao Liu, Akshat Kumar, Caiming Xiong, Silvio Savarese, and
  Doyen Sahoo.
\newblock Unified training of universal time series forecasting transformers.
\newblock In {\em Forty-first International Conference on Machine Learning},
  2024.

\bibitem{devlin2018bert}
Jacob Devlin, Ming-Wei Chang, Kenton Lee, and Kristina Toutanova.
\newblock Bert: Pre-training of deep bidirectional transformers for language
  understanding.
\newblock {\em arXiv preprint arXiv:1810.04805}, 2018.

\bibitem{liu2022scinet}
Minhao Liu, Ailing Zeng, Muxi Chen, Zhijian Xu, Qiuxia Lai, Lingna Ma, and
  Qiang Xu.
\newblock Scinet: Time series modeling and forecasting with sample convolution
  and interaction.
\newblock {\em Advances in Neural Information Processing Systems},
  35:5816--5828, 2022.

\bibitem{woo2022etsformer}
Gerald Woo, Chenghao Liu, Doyen Sahoo, Akshat Kumar, and Steven Hoi.
\newblock Etsformer: Exponential smoothing transformers for time-series
  forecasting.
\newblock {\em arXiv preprint arXiv:2202.01381}, 2022.

\bibitem{liu2021pyraformer}
Shizhan Liu, Hang Yu, Cong Liao, Jianguo Li, Weiyao Lin, Alex~X Liu, and
  Schahram Dustdar.
\newblock Pyraformer: Low-complexity pyramidal attention for long-range time
  series modeling and forecasting.
\newblock In {\em International conference on learning representations}, 2021.

\end{thebibliography}
\end{document}